\RequirePackage{fix-cm}

\documentclass[natbib,twocolumn]{svjour3}          
\smartqed  
\usepackage{graphicx}
\usepackage{xcolor}
\usepackage[pagebackref=true,breaklinks=true,citecolor=tealblue,colorlinks,linkcolor=ruby,bookmarks=false]{hyperref}
\definecolor{ruby}{rgb}{0.88, 0.07, 0.37}
\definecolor{tealblue}{rgb}{0.18, 0.40, 0.46}

\usepackage{array}
\usepackage{multirow}
\newcolumntype{H}{>{\setbox0=\hbox\bgroup}c<{\egroup}@{}}
\usepackage{amsmath}
\usepackage{amsfonts}

\usepackage{hyperref}
\usepackage{graphicx}
\usepackage{amssymb}

\usepackage{bbm}
\usepackage{dsfont}
\usepackage{pifont}
\newcommand{\xmark}{\ding{55}}
\newcommand{\tabincell}[2]{\begin{tabular}{@{}#1@{}}#2\end{tabular}}

\usepackage{microtype}

\begin{document}

\title{PV-RCNN++: Point-Voxel Feature Set Abstraction With Local Vector Representation for 3D  Object Detection
}

\author{Shaoshuai Shi \and Li Jiang \and Jiajun Deng \and Zhe Wang \and Chaoxu Guo \and Jianping Shi \and Xiaogang Wang \and Hongsheng Li
}


\institute{
S. Shi \and L. Jiang \at Max Planck Institute for Informatics, Germany\\
\email{{\{sshi, lijiang\}@mpi-inf.mpg.de}} 
\and 
J. Deng \at The University of Sydney, Australia\\
\email{{jiajun.deng@sydney.edu.au}} 
\and 
Z. Wang \and C. Guo \and J. Shi \at SenseTime Research, China\\
\email{{\{wangzhe, cxguo, shijianping\}@sensetime.com}} 
\and 
X. Wang \and H. Li \at The Chinese University of Hong Kong, Hong Kong SAR, China\\
\email{{\{xgwang, hsli\}@ee.cuhk.com.hk}} 
}

\date{Received: date / Accepted: date}

\maketitle

\begin{abstract}
3D object detection is receiving increasing attention from both industry and academia thanks to its wide applications in various fields. In this paper, we propose Point-Voxel Region-based Convolution Neural Networks (PV-RCNNs) for 3D object detection on point clouds. First, we propose a novel 3D detector, PV-RCNN, which boosts the 3D detection performance by deeply integrating the feature learning of both point-based set abstraction and voxel-based sparse convolution through two novel steps, \emph{i.e.}, the voxel-to-keypoint scene encoding and the keypoint-to-grid RoI feature abstraction. Second, we propose an advanced framework, PV-RCNN++, for more efficient and accurate 3D object detection. It consists of two major improvements: sectorized proposal-centric sampling for efficiently producing more representative keypoints, and VectorPool aggregation for better aggregating local point features with much less resource consumption. With these two strategies, our PV-RCNN++ is about $3\times$ faster than PV-RCNN, while also achieving better performance. The experiments demonstrate that our proposed PV-RCNN++ framework achieves state-of-the-art 3D detection performance on the large-scale and highly-competitive Waymo Open Dataset with 10 FPS inference speed on the detection range of $150m \times 150m$.

\keywords{
3D Object Detection \and Point Clouds\and LiDAR\and 
Autonomous Driving\and 
Sparse Convolution
}
\end{abstract}

\section{Introduction}
\label{intro}
3D object detection on point clouds aims to localize and recognize 3D objects from a set of 3D points, which is a fundamental task of 3D scene understanding and is widely-adopted in lots of real-world applications like autonomous driving, intelligent traffic system and robotics. 
Compared to 2D detection methods on images~\citep{girshick2015fast,ren2015faster,liu2016ssd,redmon2016you,lin2017feature, lin2018focal}, the sparsity and irregularity of point clouds make it challenging to directly apply 2D detection techniques to 3D detection on point clouds. 

To tackle these challenges, most of existing 3D detection methods~\citep{Chen2017CVPR,zhou2018voxelnet,yang2018pixor,yan2018second,lang2018pointpillars,chen2022mppnet} transform the points into regular voxels that can be processed with conventional 2D/3D convolutional neural networks and well-studied 2D detection heads. But the voxelization operation inevitably brings quantization errors, thus degrading their localization accuracy.
In contrast, the point-based methods~\citep{qi2017frustum,shi2019pointrcnn,wang2019frustum} naturally preserve accurate point locations in feature extraction but are generally computationally-intensive on handling large-scale points.
There are also some existing approaches~\citep{chen2019fast,li2021lidar} that simply combine these two strategies by adopting the voxel-based methods for feature extraction and 3D proposal generation in the first stage, but introducing the raw point representation in a second stage to compensate the quantization errors for fine-grained proposal refinement. 
However, this simple stacked combination ignores deep fusion of their basic operators (\emph{e.g.}, sparse convolution~\citep{3DSemanticSegmentationWithSubmanifoldSparseConvNet} and set abstraction~\citep{qi2017pointnet++}) and 
can not fully explore the feature intertwining of both strategies to take the best of both worlds.

Therefore, we propose a unified framework, namely, Point-Voxel Region-based Convolutional Neural Networks (PV-RCNNs), to take the best of both voxel and point representations by deeply integrating the feature learning strategies from both of them. The principle lies in the fact that the voxel-based strategy can more efficiently encode multi-scale features and generate high-quality 3D proposals from large-scale point clouds, while the point-based strategy can preserve accurate location information with flexible receptive fields for fine-grained proposal refinement. We demonstrate that our proposed point-voxel intertwining framework can effectively improve the 3D detection performance by deeply fusing the feature learning of both point and voxel representations.

Firstly, we introduce our initial 3D detection framework, PV-RCNN, which is a two-stage 3D detector on point clouds.
It consists of two novel steps for point-voxel feature aggregation.
The first step is voxel-to-keypoint scene encoding, where a 3D voxel CNN with sparse convolutions is adopted for feature learning and proposal generation. The multi-scale voxel features are then summarized into a small set of keypoints by point-based set abstraction, where the keypoints with accurate point locations are sampled by farthest point sampling from the raw points. 
The second step is keypoint-to-grid RoI feature abstraction, where we propose RoI-grid pooling module to aggregate the above keypoint features back to regular RoI grids of each proposal. It encodes multi-scale contextual information to form regular grid features for proposal refinement.
These two steps establish feature intertwining between point-based set abstraction and voxel-based sparse convolution, which have been experimentally evidenced to improve the model representative ability as well as the detection performance.

Secondly, we propose an advanced two-stage detection network, PV-RCNN++, on top of PV-RCNN, for achieving more accurate, efficient and practical 3D object detection. The improvements of PV-RCNN++ lie in two aspects.
The first aspect is a novel sectorized proposal-centric keypoint sampling strategy, where we concentrate the limited number of keypoints in and around the 3D proposals to encode more effective scene features. 
Meanwhile, by considering radial distribution of LiDAR points, we propose to conduct point sampling parallelly in different sectors, which accelerates keypoint sampling process, while also ensuring uniform distribution of keypoints.
Our proposed keypoint sampling strategy is much faster and more effective than vanilla farthest point sampling that has a quadratic complexity. The efficiency of the whole framework is thus greatly improved, which is particularly important for large-scale 3D scenes with millions of points. 
The second aspect is a novel local feature aggregation module, VectorPool aggregation, for more effective and efficient local feature encoding on point clouds.
We argue that the relative point locations in a local region are robust, effective and discriminative features for describing local geometry.
We propose to split 3D local space into regular and compact sub-voxels, the features of which are sequentially concatenated to form a hyper feature vector.  
The sub-voxel features in different locations are encoded with separate kernel weights to generate position-sensitive local features. 
In this way, different local location information is encoded with different feature channels in the hyper feature vector.
Compared with set abstraction, our VectorPool aggregation can efficiently handle a very large number of centric points due to the compact local feature representation.
Equipped with VectorPool aggregation in both voxel-based backbone and RoI-grid pooling module, our PV-RCNN++ is more memory-friendly and faster than previous counterparts with comparable or even better performance, which helps in establishing a practical 3D detector for resource-limited devices.

In a nutshell, our contributions are three-fold:
1) Our PV-RCNN adopts two novel strategies, voxel-to-keypoint scene encoding and keypoint-to-grid RoI feature abstraction, to deeply integrate the advantages of both point-based and voxel-based feature learning strategies. 
2) Our PV-RCNN++ takes a step in more practical 3D detection system with better performance, less resource consumption and faster running speed. This is enabled by our proposed sectorized proposal-centric keypoint sampling to obtain more representative keypoints with faster speed, and is also powered by our novel VectorPool aggregation for achieving local aggregation on a large number of central points with less resource consumption and more effective representation. 
(3) Our proposed 3D detectors surpass all published methods with remarkable margins on the challenging large-scale Waymo Open Dataset.
In particular, our PV-RCNN++ achieves state-of-the-art results with 10 FPS inference speed for $150m \times 150m$ detection range.
The source code is available at \url{https://github.com/open-mmlab/OpenPCDet}.

\section{Related Work}
\noindent
\textbf{3D Object Detection with 2D images.} Image-based 3D detection aims to estimate 3D bounding boxes from a monocular image or stereo images.
Mono3D \citep{chen2016monocular} generates 3D proposals with ground-plane assumption, which are scored by exploiting semantic knowledge from images. 
The following works \citep{mousavian20173d,li2019gs3d} incorporate the relations between 2D and 3D boxes as geometric constraint. M3D-RPN \citep{brazil2019m3d} introduces a 3D region proposal network with depth-aware convolutions. \citep{chabot2017deep,murthy2017reconstructing,manhardt2019roi} predict 3D boxes based on a wire-frame template obtained from CAD models. 
RTM3D \citep{li2020rtm3d} 
performs coarse keypoints detection to localize 3d objects.
On the stereo side, Stereo R-CNN \citep{li2019stereo,qian2020end} capitalizes on a stereo RPN to associate proposals from left and right images. 
DSGN \citep{chen2020dsgn} introduces differentiable 3D volume to learn depth information and semantic cues in an end-to-end optimized pipelines. 
LIGA-Stereo \citep{guo2021liga} proposes to learn good geometry features from the well-trained LiDAR detector. 
Pseudo-LiDAR based approaches \citep{wang2019pseudo,qian2020end,you2020pseudo} covert the image pixels to artificial point clouds, where the LiDAR-based detectors can operate on them for 3D box estimation. 
These image-based 3D detection methods suffer from inaccurate depth estimation and can only generate coarse 3D bounding boxes.

Recently, in addition to image-based 3D detection from monocular image or stereo images, a comprehensive scene understanding with surrounding cameras has drawn a lot of attention, where the well-known bird's-eye-view (BEV) representation is generally adopted for better feature fusion from multiple surrounding images. 
LSS~\citep{philion2020lift} and CaDDN~\citep{reading2021categorical} predicts depth distribution to ``lift'' the 2D image features to a BEV feature map for 3D detection. Their follow-up works~\citep{huang2021bevdet,huang2022bevdet4d,li2022bevdepth,xie2022m} learn a depth-based implicit projection  to project image features to BEV space.
Some other works also explore transformer structure to project image features from perspective view to BEV space via cross attention, such as DETR3D~\citep{wang2022detr3d}, PETR~\citep{liu2022petr,liu2022petrv2}, BEVFormer~\citep{li2022bevformer}, PolarFormer~\citep{jiang2022polarformer}, etc. 
Although these works greatly improve the performance of image-based 3D detection by projecting multi-view images to a unified BEV space, the inaccurate depth estimation is still the main challenge for image-based 3D detection.

\noindent
\textbf{Representation Learning on Point Clouds.}
Recently representation learning on point clouds has drawn lots of attention for improving the performance of 3D classification and segmentation \citep{qi2017pointnet,qi2017pointnet++,wang2019dynamic,huang2018recurrent,zhao2019pointweb,li2018pointcnn,su2018splatnet,wu2019pointconv,jaritz2019multi,jiang2019hierarchical,thomas2019kpconv,choy20194d,liu2020closer}. 
In terms of 3D detection, previous methods generally project the points to regular 2D pixels \citep{Chen2017CVPR,yang2018pixor} or 3D voxels \citep{zhou2018voxelnet,chen2019fast} for processing them with 2D/3D CNN. Sparse convolution \citep{3DSemanticSegmentationWithSubmanifoldSparseConvNet} is adopted in \citep{yan2018second,shi2020part} to effectively learn sparse voxel features from point clouds. Qi et al. \citep{qi2017pointnet,qi2017pointnet++} proposes PointNet to directly learn point features from raw points, where set abstraction enables flexible receptive fields by setting different search radii. \citep{liu2019point} combines both voxel CNN and point multi-layer percetron network for efficient point feature learning. 
In comparison, our PV-RCNNs take advantages from both voxel-based (\emph{i.e.}, 3D sparse convolution) and point-based (\emph{i.e.}, set abstraction) strategies to enable both high-quality 3D proposal generation with dense BEV detection heads and flexible receptive fields in 3D space for improving 3D detection performance. 

\noindent
\textbf{3D Object Detection with Point Clouds.}
Most of existing 3D detection approaches can be roughly classified into three categories in terms of different strategies to learn point cloud features, \emph{i.e.}, the voxel-based methods, the point-based methods as well as the methods combining both points and voxels. 

The voxel-based methods project point clouds to regular grids to tackle the irregular data format problem. 
MV3D \citep{Chen2017CVPR} projects points to 2D bird view grids and places lots of predefined 3D anchors for generating 3D boxes, and the following works \citep{ku2018joint,Liang2018ECCV,Liang2019CVPR,vora2020pointpainting,yoo20203d,huang2020epnet} develop better strategies for multi-sensor fusion. \citep{yang2018pixor,Yang2018CoRL,lang2018pointpillars} introduce more efficient frameworks with bird-eye view representation while \citep{ye2020hvnet} proposes to fuse grid features of multiple scales. 
MVF \citep{zhou2020end} integrates 2D features from bird-eye view and perspective view before projecting points into pillar representations \citep{lang2018pointpillars}. 
Some other works \citep{song2016deep,zhou2018voxelnet,wang2022cagroup3d} divide the points into 3D voxels to be processed by 3D CNN. 3D sparse convolution \citep{3DSemanticSegmentationWithSubmanifoldSparseConvNet} is introduced by \citep{yan2018second} for efficient 3D voxel processing.
\citep{wang2019voxelFPN} utilizes multiple detection heads for detecting 3D objects with different scales. In addition, \citep{wang2020pillar,chen2020hotspots} predicts bounding box parameters following anchor-free paradigm. These grid-based methods are generally efficient for accurate 3D proposal generation but the receptive fields are constraint by the kernel size of 2D/3D convolutions.

\begin{figure*}
	\begin{center}
		\includegraphics[width=1.0\linewidth,height=6.2cm]{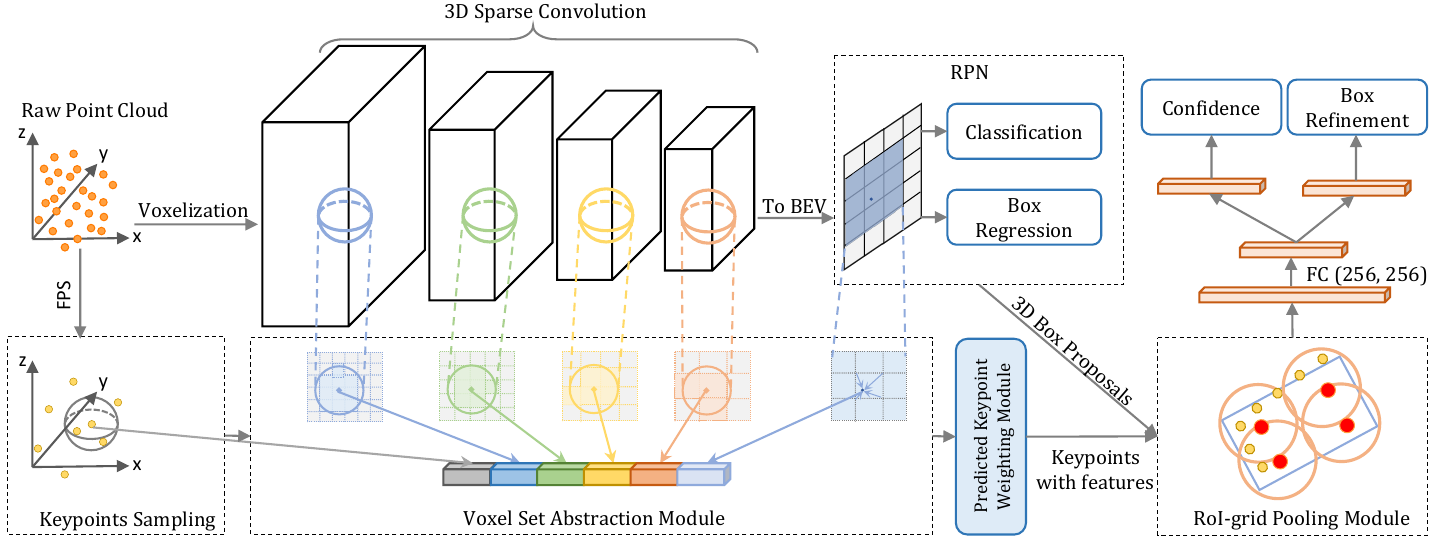}
	\end{center}
	\caption{The overall architecture of our proposed PV-RCNN. The raw point clouds are first voxelized to feed into the 3D sparse convolution based encoder to learn multi-scale semantic features and generate 3D object proposals. Then the learned voxel-wise feature volumes at multiple neural layers are summarized into a small set of key points via the novel voxel set abstraction module. Finally the keypoint features are aggregated to the RoI-grid points to learn proposal specific features for fine-grained proposal refinement and confidence prediction.}
	\label{fig:framework_v1}
\end{figure*}

The point-based methods directly detect 3D objects from raw points. 
F-PointNet \citep{qi2017frustum} applies PointNet \citep{qi2017pointnet,qi2017pointnet++} for 3D detection from the cropped points based on 2D image boxes. PointRCNN \citep{shi2019pointrcnn} generates 3D proposals directly from raw points by only taking 3D points, and some  works~\citep{qi2019deep,cheng2021back,wang2022rbgnet} follow the point-based pipeline by exploring different feature grouping strategy for generating 3D boxes. 3DSSD \citep{yang20203dssd} introduces hybrid feature-distance based farthest point sampling on raw points.
These point-based methods are mostly based on PointNet series, especially set abstraction \citep{qi2017pointnet++}, which enables flexible receptive fields for point cloud feature learning. However, it is challenging to extend these point-based methods to large-scale point clouds since they generally consume much more memory/computation resources than the above voxel-based methods.

There are also some works that utilize both the point-based and voxel-based representations. STD \citep{std2019yang} transforms point-wise features to dense voxels for refining the proposals. Fast Point R-CNN~\citep{chen2019fast} fuses the deep voxel features with raw points for 3D detection. Part-A2-Net~\citep{shi2020part} aggregates the point-wise part locations by the voxel-based RoI-aware pooling to improve 3D detection performance. 
However, these methods generally simply transform features between two representations and do not fuse the deeper features from the specific basic operations of these two representations.
In contrast, our PV-RCNN frameworks explore on how to deeply aggregate features by learning with both point-based (\emph{i.e.}, set abstraction) and voxel-based (\emph{i.e.}, sparse convolution) feature learning modules to boost the 3D detection performance.

\section{PV-RCNN: Point-Voxel Feature Set Abstraction for 3D Object Detection}\label{sec:pvrcnnv1}
Most of state-of-the-art 3D detectors~\citep{shi2020part,yin2021center,sheng2021} adopt 3D sparse convolution for learning representative features from irregular points thanks to its efficiency and effectiveness on handling large-scale point clouds. However, 3D sparse convolution network suffers from losing accurate point information due to the indispensable voxelization process. 
In contrast, the point-based approaches~\citep{qi2017pointnet,qi2017pointnet++}  naturally preserve accurate point locations and can capture rich context information with flexible receptive fields, where the accurate point locations are essential for estimating accurate 3D bounding boxes.

In this section, we briefly review our initial 3D detection framework, PV-RCNN~\citep{shi2020pv}, for 3D object detection from point clouds. It deeply integrates the voxel-based sparse convolution and point-based set abstraction operations to take the best of both worlds.

As shown in Fig.~\ref{fig:framework_v1}, PV-RCNN is a two-stage 3D detection framework that adopts a 3D voxel CNN with sparse convolution as the backbone for efficient feature encoding and proposal generation (Sec.~\ref{sec:rpn}), and then we generate the proposal-aligned features for predicting accurate 3D bounding boxes by intertwining point-voxel features through two novel steps, which are voxel-to-keypoint scene encoding (Sec.~\ref{sec:vsa}) and keypoint-to-grid RoI feature abstraction (Sec.~\ref{sec:roi_grid_pool}).  

\subsection{Voxel Feature Encoding and Proposal Generation}\label{sec:rpn}
In order to handle 3D object detection on the large-scale point clouds, we adopt the 3D voxel CNN with sparse convolution~\citep{3DSemanticSegmentationWithSubmanifoldSparseConvNet} as the backbone network to generate initial 3D proposals.

The input points $\mathcal{P}$ are first divided into small voxels with spatial resolution of $L\times W \times H$, where non-empty voxel features are directly calculated by averaging the coordinates of inside points. 
The network utilizes a series of 
3D sparse convolution to gradually convert the points into feature volumes with $1\times, 2\times$, $4\times$, $8\times$ downsampled sizes. 
We follow \citep{yan2018second} to stack the 3D feature volumes along $Z$ axis to obtain the $\frac{L}{8}\times\frac{W}{8}$ bird-view feature maps, which can be naturally combined with the 2D detection heads~\citep{liu2016ssd,yin2021center} for high quality 3D proposal generation.  

It is worth noting that the sparse feature volumes at each level can be viewed as a set of sparse voxel-wise feature vectors, and these multi-scale semantic features are considered as the input of our following voxel-to-keypoint scene encoding step.

\subsection{Voxel-to-Keypoint Scene Encoding}\label{sec:vsa}
Given the multi-scale scene features, we propose to summarize these features into a small number of keypoints, which serve as the courier to propagate features from the above 3D voxel CNN to the refinement network.

\noindent
\textbf{Keypoint Sampling.} 
We simply adopt farthest point sampling algorithm as in~\citep{qi2017pointnet++} to sample a small number of keypoints 
$\mathcal{K}=\left\{p_i\mid p_i \in \mathbb{R}^3\right\}_{i=1}^{n}$ 
from the raw points $\mathcal{P}$, where $n$ is 
a hyper-parameter 
(\emph{e.g.}, n=4,096 for Waymo Open Dataset~\citep{Sun_2020_CVPR}).
It encourages that the keypoints are uniformly distributed around non-empty voxels and can be representative to the overall scene.

\noindent
\textbf{Voxel Set Abstraction Module.}
To aggregate the multi-scale semantic features from 3D feature volumes to the keypoints, we propose \textit{Voxel Set Abstraction} (VSA) module.
The set abstraction \citep{qi2017pointnet++} is adopted for aggregating voxel-wise feature volumes. 
The key difference is that the surrounding local points are now regular voxel-wise semantic features from 3D voxel CNN, instead of the neighboring raw points with features learned by PointNet~\citep{qi2017pointnet}.

Specifically, we denote the number of non-empty voxels in the $k$-th level of 3D voxel CNN as $N_k$, and the voxel-wise features and 3D coordinates are denoted as 
$\mathcal{F}^{(l_k)}=\left\{[f_i^{(l_k)}, v_{i}^{(l_k)}]\mid f_i^{(l_k)}\in \mathbb{R}^{C}, v_{i}^{(l_k)} \in \mathbb{R}^{3} \right\}_{i=1}^{N_k}$, where $C$ indicates the number of feature dimensions.

For each keypoint $p_i\in \mathcal{K}$, to retrieve the set of neighboring voxel-wise feature vectors, we first identify its neighboring non-empty voxels at the $k$-th level within a radius $r_k$ as
\begin{align}
	S_i^{(l_k)} =\left\lbrace \left[f_j^{(l_k)}, v_j^{(l_k)} - p_i\right] \;\middle|\;
	\begin{tabular}{@{}l@{}}
		$\left\Vert v_j^{(l_k)} - p_i \right\Vert < r_k$\\
	\end{tabular}
	\right\rbrace,
\end{align}
where $[f_j^{(l_k)}, v_{j}^{(l_k)}] \in \mathcal{F}^{(l_k)}$, 
and the local relative position $v_j^{(l_k)}$ $- p_i$ is concatenated to indicate the relative location of $f_j^{(l_k)}$ in this local area.
The features within neighboring set $S_i^{(l_k)}$ are then aggregated by a PointNet-block~\citep{qi2017pointnet} to generate keypoint  feature as
\begin{align}\label{eq:pointnet}
		f_i^{(\text{pv}_k)} = \max \left\{\text{SharedMLP}\left(S_i^{(l_k)}\right)\right\},
\end{align}

\noindent 
where $\text{SharedMLP}(\cdot)$ denotes a shared multi-layer perceptron (MLP) network to encode voxel-wise features and relative locations, and $\max\{\cdot\}$ conducts permutation invariant feature aggregation to 
map diverse number of neighboring voxel features to a single keypoint feature $f_i^{(\text{pv}_k)}$.
Here multiple radii are utilized to capture richer contextual information. 

The above voxel feature aggregation is performed at the outputs of different levels of 3D voxel CNN, and the aggregated features from different scales are concatenated to obtain the multi-scale semantic feature for keypoint $p_i$ as 
\begin{align}\label{eq:keypointfeature0}
	f_i^{(p)} = \text{Concat}\left(\left\{f_i^{(\text{pv}_k)}\right\}_{k=1}^{4}, f_i^{(\text{raw})}, f_i^{(\text{bev})} \right), 
\end{align}
where $i \in \{1, \dots, n\}$, and $k \in \{1, \dots, 4\}$ indicates that the keypoint features are aggregated from four-level voxel-wise features  of 3D voxel CNN. 
Note that the keypoint features are further enriched with two extra information sources, where the raw point features $f_i^{(\text{raw})}$ are aggregated as in Eq.~\eqref{eq:pointnet} to partially make up the quantization loss of point voxelization, while 2D bird-view features $f_i^{(\text{bev})}$ are obtained by bilinear interpolation on the $8\times$ downsampled 2D feature maps to achieve larger receptive fields along the height axis.

\begin{figure}[t]
	\begin{center}
		\includegraphics[width=0.95\linewidth]{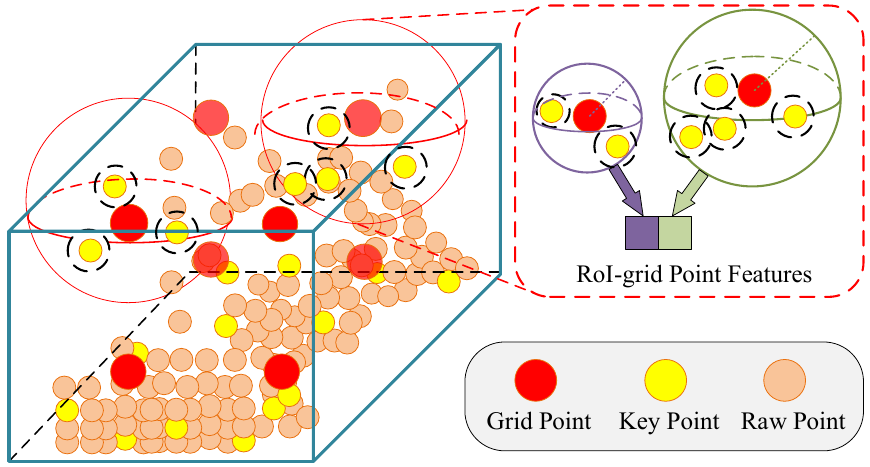}
	\end{center}
	\vspace{-0.2cm}
	\caption{Illustration of RoI-grid pooling module. Rich context information of each 3D RoI is aggregated by set abstraction operation with multiple receptive fields.}
	\label{fig:roipool}
\end{figure}

\noindent
\textbf{Predicted Keypoint Weighting.}~
Intuitively, the keypoints belonging to the foreground objects should contribute more to the proposal refinement, while the keypoints from the background regions should contribute less.
Hence, we propose a \textit{Predicted Keypoint Weighting} (PKW) module to re-weight the keypoint features with extra supervisions from point segmentation as 
\begin{align}\label{eq:pkw}
	f_i^{(p)} = \text{MLP}(f_i^{(p)}) \cdot f_i^{(p)},
\end{align}
where $\text{MLP}(\cdot)$ is a three-layer MLP network with a sigmoid function to predict foreground confidence. 
It is trained with focal loss \citep{lin2018focal} by the default parameters and the segmentation labels can be directly generated from the 3D box annotations as in ~\citep{shi2019pointrcnn}. 
Note that this PKW module is optional for our framework as it only leads small gains (see Table.~\ref{tab:exp_pkw}).

The keypoint features $\mathcal{F}=\left\{f^{(p)}_i\right\}_{i=1}^n$ not only incorporate multi-scale semantic features from the 3D voxel backbone network, but also naturally preserve accurate location information through its 3D keypoint coordinates $\mathcal{K}=\left\{p_i\right\}_{i=1}^n$, which provides strong capacity of preserving 3D structural information of the entire scene for the following fine-grained proposal refinement.

\subsection{Keypoint-to-Grid RoI Feature Abstraction}\label{sec:roi_grid_pool}
Given the aggregated keypoint features and their 3d coordinates, in this step, we propose keypoint-to-grid RoI feature abstraction to generate accurate proposal-aligned features for fine-grained proposal refinement. 

\noindent
\textbf{RoI-grid Pooling via Set Abstraction.}~
We propose the \textit{RoI-grid pooling} module to aggregate the keypoint features to the RoI-grid points by adopting multi-scale local feature grouping.
For each given 3D proposal, we uniformly sample $6\times 6\times 6$ grid points according to the 3D proposal box, which are then flattened and denoted as $\mathcal{G}=\{g_i\}_{i=1}^{6\times 6\times 6=216}$.
To aggregate the features of keypoints to the RoI grid points, 
we firstly identify the neighboring keypoints of a grid point $g_i$ as 
\begin{align}
	\Psi =\left\lbrace \left[f_j^{(p)}, p_j - g_i\right] \;\middle|\;
	\begin{tabular}{@{}l@{}}
		$\left\Vert p_j - g_i \right\Vert < r^{(g)}$\\
	\end{tabular}
	\right\rbrace,
\end{align}
where $p_j \in \mathcal{K}$ and $f_j^{(p)} \in \mathcal{F}$. We concatenate $p_j-g_i$ to indicate the local relative location within the ball of radius $r^{(g)}$. 
Then we adopt the similar process with Eq.~\eqref{eq:pointnet} to summarize the neighboring keypoint feature set $\Psi$ to obtain the features of grid point $g_i$ as 
\begin{align}\label{pointnet2}
	f_i^{(g)} = \max \left\{\text{SharedMLP}\left(\Psi\right)\right\}.
\end{align}

Note that we set multiple radii $r^{(g)}$ and aggregate keypoint features with different receptive fields, which are concatenated together for capturing richer multi-scale contextual information. 
Next, all RoI-grid features $\{f_i^{(g)}\}_{i=1}^{216}$ of the same RoI can be vectorized and transformed by a two-layer MLP with $256$ feature dimensions to represent the overall features of this proposal box. 

Our proposed RoI-grid pooling operation can aggregate much richer contextual information than the previous RoI-pooling/RoI-align operation \citep{shi2019pointrcnn,std2019yang,shi2020part}. 
It is because a single keypoint can contribute to multiple RoI-grid points due to the overlapped neighboring balls of RoI-grid points, and their receptive fields are even beyond the RoI boundaries by capturing the contextual keypoint features outside the 3D RoI. In contrast, the previous state-of-the-art methods either simply average all point-wise features within the proposal as the RoI feature \citep{shi2019pointrcnn}, or pool many uninformative zeros as the RoI features because of the very sparse point-wise features \citep{shi2020part,std2019yang}.

\noindent
\textbf{Proposal Refinement.} 
Given the above RoI-aligned features, the refinement network learns to predict the size and location (\emph{i.e.} center, size and orientation) residuals relative to the 3D proposal box. 
Two sibling sub-networks are employed for confidence prediction and proposal refinement. Each sub-network consists of a two-layer MLP and a linear prediction layer. 
We follow \citep{shi2020part} to conduct the IoU-based confidence prediction.
The binary cross-entropy loss is adopted to optimize the IoU branch while the box residuals are optimized with smooth-L1 loss.

\begin{figure*}
	\begin{center}	\includegraphics[width=0.99\linewidth]{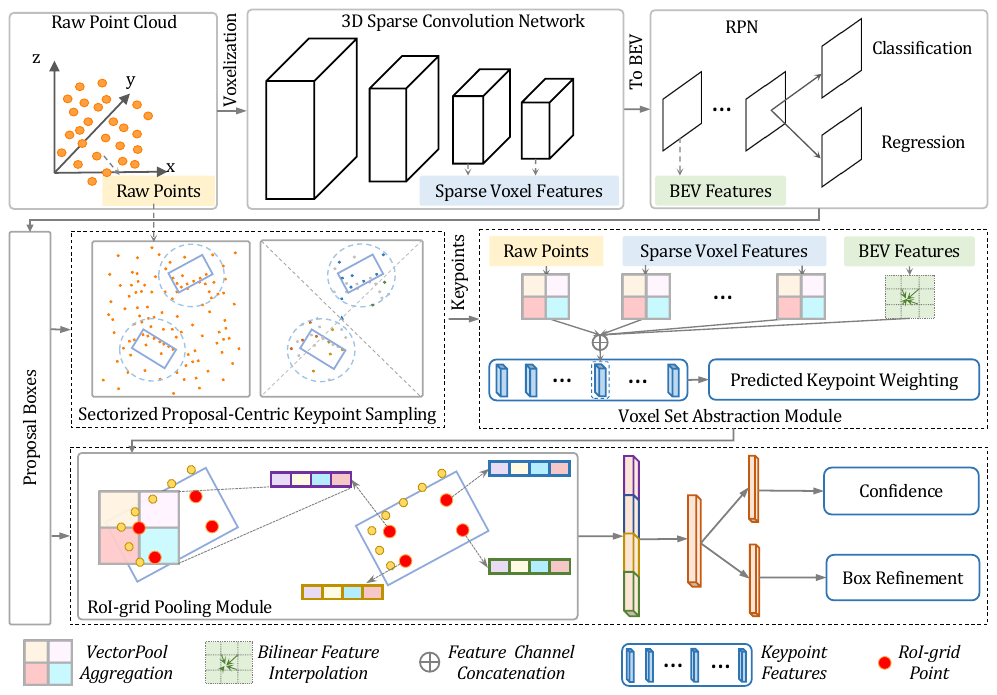}
	\end{center}
	\vspace{-2mm}
	\caption{The overall architecture of our proposed PV-RCNN++ framework. We propose sectorized proposal-centric keypoint sampling module to concentrate keypoints to the neighborhoods of 3D proposals while it can also accelerate the process with sectorized farthest point sampling. Moreover, our proposed VectorPool module is utilized in both voxel set abstraction module and RoI-grid pooling module to improve the local feature aggregation and save memory/computation resources.}
	\label{fig:framework_v2}
\end{figure*}

\section{PV-RCNN++: Faster and Better 3D Detection With PV-RCNN Framework}
Although our proposed PV-RCNN 3D detection framework achieves state-of-the-art performance~\citep{shi2020pv}, it suffers from the efficiency problem when handling large-scale point clouds. 
To make PV-RCNN framework more practical for real-world applications, we propose an advanced 3D detection framework, \emph{i.e.,} PV-RCNN++, for more accurate and efficient 3D object detection with less resource consumption. 

As shown in Fig.~\ref{fig:framework_v2}, we present two novel modules to improve both the accuracy and efficiency of PV-RCNN framework. One is sectorized proposal-centric strategy for much faster and better keypoint sampling, and the other one is VectorPool aggregation module for more effective and efficient local feature aggregation from large-scale point clouds. 
These two modules are adopted to replace their counterparts in PV-RCNN, which are introduced in Sec.~\ref{sec:spc} and Sec.~\ref{sec:vectorpool}, respectively.

\subsection{Sectorized Proposal-Centric Sampling for Efficient and Representative Keypoint Sampling}\label{sec:spc}
The keypoint sampling is critical for PV-RCNN framework as keypoints bridge the point-voxel representations and heavily influence the performance of proposal refinement. 
However, previous keypoint sampling algorithm (see Sec.~\ref{sec:vsa}) has two main drawbacks. 
(i) Farthest point sampling is time-consuming due to its quadratic complexity, which hinders the training and inference speed of PV-RCNN, especially for keypoint sampling on large-scale point clouds.
(ii) 
It would generate a large number of background keypoints that are generally useless to proposal refinement, since only the keypoints around proposals can be retrieved by RoI-grid pooling module. 

To mitigate these drawbacks, we propose the \textit{Sectorized Proposal-Centric} (SPC) keypoint sampling to uniformly sample keypoints from more concentrated neighboring regions of proposals, while also being much faster than the vanilla farthest point sampling algorithm. 
It mainly consists of two novel steps, which are the proposal-centric filtering and the sectorized sampling, which are illustrated in the following paragraphs.

\begin{figure*}
	\begin{center}
		\includegraphics[width=1.0\linewidth]{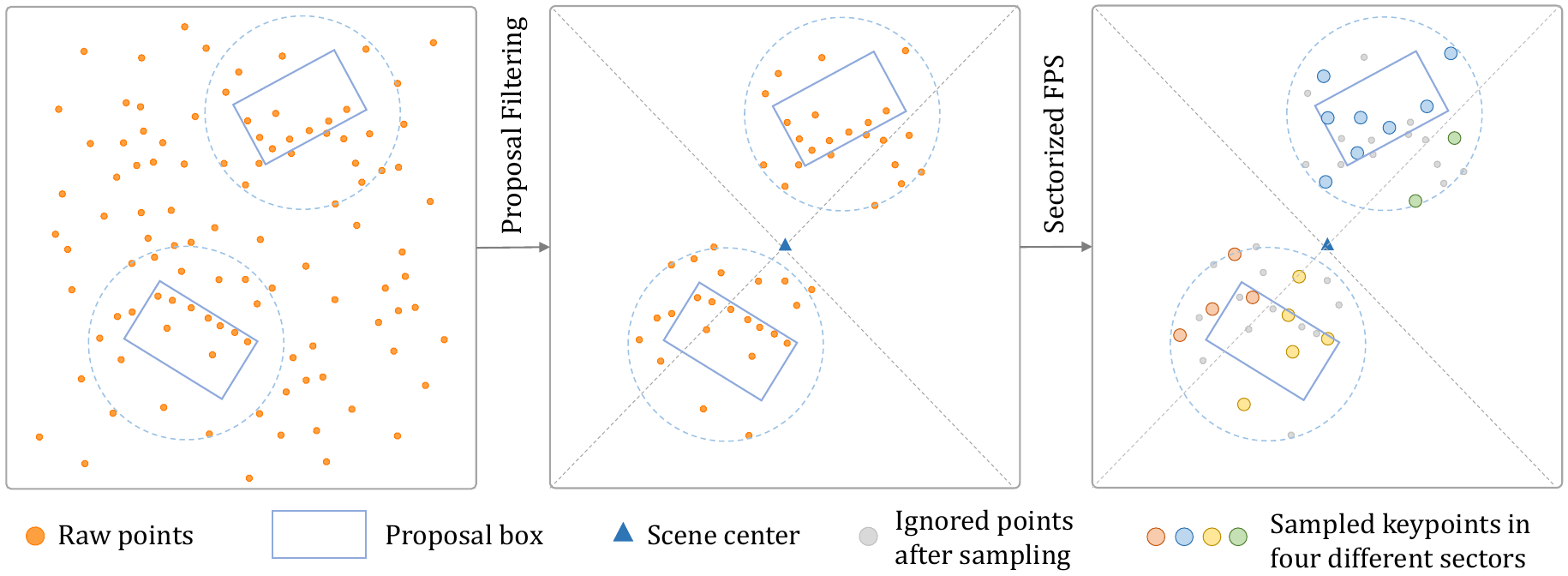}
	\end{center}
	\vspace{-4mm}
	\caption{Illustration of Sectorized Proposal-Centric (SPC) keypoint sampling. It contains two steps, where the first proposal filtering step concentrates the limited number of keypoints to the neighborhoods of proposals, and the following sectorized-FPS step divides the whole scene into several sectors for accelerating the keypoint sampling process while also keeping the keypoints uniformly distributed. }
	\label{fig:spc_draw}
\end{figure*}

\noindent
\textbf{Proposal-Centric Filtering.}
To better concentrate the keypoints on the more important areas and also reduce the complexity of the next sampling process, we first adopt the proposal-centric filtering step. 

Specifically, we denote a number of $N_p$ 3D proposals as $\mathcal{D}=\{[c_i, d_i] \mid c_i \in \mathbb{R}^3, d_i \in \mathbb{R}^3\}_{i=1}^{N_p}$, where $c_i$ and $d_i$ are the center and size of each proposal box, respectively. 
We restrict the keypoint candidates $\mathcal{P}'$ to the neighboring point sets of all proposals as
\begin{align}
	\mathcal{P}' =\left\lbrace p_i \;\middle|\;
	\begin{tabular}{@{}c@{}}
		$\left\Vert p_i - c_j \right\Vert <  \frac{1}{2} \cdot {\max \left(d_j\right)} + r^{(s)}$ \\
	\end{tabular}
	\right\rbrace,
\end{align}
where $[c_j, d_j] \in \mathcal{D}$, $p_i \in \mathcal{P}$ indicates the raw point, and $\max(\cdot)$ obtains the maximum length of 3D box size. 
$r^{(s)}$ is a hyperparameter indicating the maximum extended radius of the proposals.
Through this proposal-centric filtering process, the number of candidate keypoints for sampling is greatly reduced from $|\mathcal{P}|$ to $|\mathcal{P}'|$. For instance, for the Waymo Open Dataset~\citep{Sun_2020_CVPR}, generally $\mathcal{P}$ is about $180k$ and $\mathcal{P}'$ can be smaller than $90k$ in most cases (the exact point number depends on the number of proposal boxes in each scene).

Hence, this step not only reduces the time complexity of the follow-up keypoint sampling, but also concentrates the limited number of keypoints to better encode the neighboring regions of the proposals.

\noindent
\textbf{Sectorized Keypoint Sampling.}
To further parallelize the keypoint sampling process for acceleration, as shown in Fig.~\ref{fig:spc_draw}, we propose the sectorized keypoint sampling strategy, which takes advantage of radial distribution of the LiDAR points to better parallelize and accelerate the keypoint sampling process. 

Specifically, we divide proposal-centric point set $\mathcal{P}'$ into $s$ sectors centered at the scene center, and the point set of $k$-th sector can be represented as
\begin{align}
	S'_k =\left\lbrace p_i \;\middle|\;
	\begin{tabular}{@{}c@{}}
		$\left\lfloor \left(\text{arctan}\left(p_i^{y}, p_i^{x}\right) + \pi\right) \cdot \frac{s}{2\pi} \right\rfloor = k - 1$
	\end{tabular}
	\right\rbrace,
\end{align}
where $k \in \{1, \dots, s\}$, $p_i=(p_i^x, p_i^y, p_i^z) \in \mathcal{P}'$
, and $\text{arctan}($ $p_i^y, p_i^x)$ $\in (-\pi, \pi]$ indicates the angle between the positive $X$ axis and the ray ended with $(p_i^x, p_i^y)$ in terms of the bird's eye view.

Through this process, we divide the task of sampling $n$ keypoints into $s$ subtasks of sampling local keypoints, where $k$-th sector samples $\left\lfloor\frac{|S'_k|}{|P'|} \times n \right\rfloor$ keypoints from the point set $S'_k$. These subtasks are eligible to be executed in parallel on GPUs, while the scale of keypoint sampling (\emph{i.e.}, time complexity) is further reduced from $|\mathcal{P}'|$ to $\max_{k \in \{1, \dots, s\}} |S'_k|$.
Note that we adopt farthest point sampling in each subtask since both the qualitative and quantitative experiments in Sec.~\ref{sec:ab_study} demonstrate that farthest point sampling can  generate more uniformly distributed keypoints to better cover the whole regions, which is critical for the final detection performance.

It is worth noting that our sector-based group partition can roughly produce similar number of points in each group by considering radial distribution of the points generated by LiDAR sensors, which is essential to speed up the keypoint sampling since the overall running time depends on the group with the most points.

Therefore, our proposed keypoint sampling algorithm greatly reduces the scale of keypoint sampling from $|\mathcal{P}|$ to the much smaller $\max_{k \in \{1, \dots, s\}} |S'_k|$, which not only effectively accelerates the keypoint sampling process, but also increases the capability of keypoint feature representation by concentrating the keypoints to the more important neighboring regions of 3D proposals.

Although the proposed sectorized keypoint sampling is tailored for LiDAR sensors, the main idea behind it, that is, conducting
FPS in spatial groups to speed up the operation, is also effective with other types of sensors.  It should be noted that the point group generation should be based on
spatially partitioning to keep the overall uniform distribution. As shown in Table~\ref{tab:exp_spc}, randomly dividing the points
into groups, while ensuring a balance in the number of points between groups, harms the model
performance.

\begin{figure*}
	\begin{center}
		\includegraphics[width=1.0\linewidth]{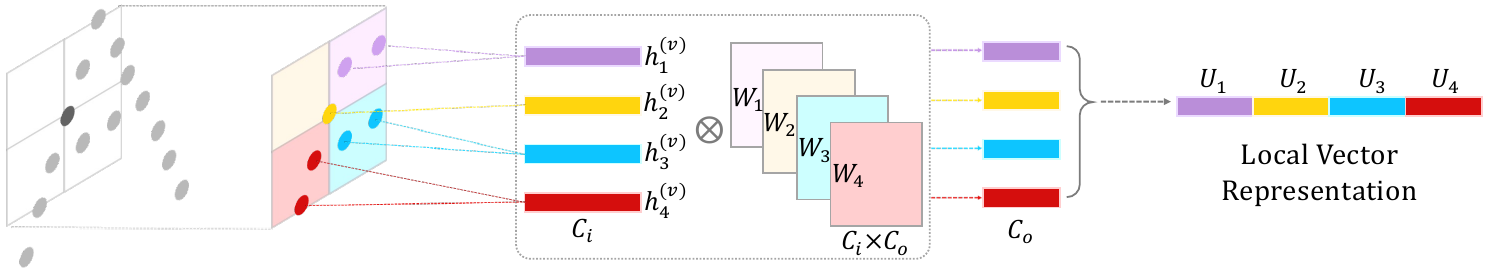}
	\end{center}
	\vspace{-2mm}
	\caption{Illustration of VectorPool aggregation for local feature aggregation from point clouds. The local 3D space around a center point is divided into dense sub-voxels, where the inside point-wise features are generated by interpolating from three nearest neighbors. 
		The features of each volume are encoded with position-specific kernels to generate position-sensitive features, which are sequentially concatenated to generate the local vector representation to explicitly encode the spatial structure information.
		Note that the notations in the figure follows the same denition as in Sec.~\ref{sec:vectorpool}, except that we simplify the channel definition of the kernels as: $C_i=9+C_{\text{in}}$, $C_o=C_{\text{mid}}$.
	}
	\label{fig:vectorpool}
\end{figure*}

\subsection{Local Vector Representation for Structure-Preserved Local Feature Learning}
\label{sec:vectorpool}
The local feature aggregation of point clouds plays an important role in PV-RCNN framework as it is the fundamental operation to deeply integrate the point-voxel features in both voxel set abstraction and RoI-grid pooling modules. 
However, we observe that set abstraction (see Eqs.~\eqref{eq:pointnet} and \eqref{pointnet2}) in PV-RCNN framework 
can be extremely time- and resource-consuming on large-scale point clouds, since it applies several shared-parameter MLP layers on the point-wise features of each local point separately. 
Moreover, the max-pooling operation in set abstraction abandons the spatial distribution information of local points and harms the representation capability of locally aggregated features from point clouds.

Therefore, in this section, we propose VectorPool aggregation module for local feature aggregation on the large-scale point clouds, 
which can better preserve spatial point distribution of local neighborhoods and also costs less memory/computation resources than the commonly-used set abstraction. 
Our PV-RCNN++ framework adopts it as a basic module to enable more effective and efficient 3D object detection.

\noindent
\textbf{Problem Statement.}~
The VectorPool aggregation module aims to generate the informative local features for $N$ target center points (denoted as $\mathcal{Q}=\{q_k \mid q_k \in \mathbb{R}^3\}_{k=1}^{N}$) by learning from $M$ given support points and their features (denoted as $\mathcal{I}=\{[h_i, a_i] \mid h_i \in \mathbb{R}^{C_{\text{in}}},$ $a_i \in \mathbb{R}^3\}_{i=1}^M$), where $C_{\text{in}}$  is the input feature channels and we are going to extract $N$ local point-wise features with $C_{\text{out}}$ channels for each point in $\mathcal{Q}$.

\noindent
\textbf{VectorPool Aggregation on Point Clouds.}~
In our proposed VectorPool aggregation module, we propose to generate position-sensitive local features by encoding different spatial regions  with separate kernel weights and separate feature channels, which are then concatenated as a single vector representation to explicitly represent the spatial structures of local point features. 

Specifically, given a target center point $q_k$, we first identify the support points that are within its cubic neighboring region, which can be represented as  
\begin{align}
	\mathcal{Y}_k =\left\lbrace \left[h_j, a_j\right] \;\middle|\;
	\begin{tabular}{@{}l@{}}
		$\max (a_j - q_k) < 2\times \delta$\\
	\end{tabular}
	\right\rbrace,
\end{align}
where $[h_j, a_j] \in \mathcal{I}$, $\delta$ is the half length of this cubic space, and $\max(a_j-q_k) \in \mathbb{R}$ obtains the maximum axis-aligned value of this 3D distance. 
Note that we double the half length (\emph{e.g.,} $2\times \delta$) of the original cubic space to contain more neighboring points for local feature aggregation of this target point. 

To generate position-sensitive features for this local cubic neighborhood centered at $q_k$, we split its neighboring cubic space 
into $n_x\times n_y\times n_z$ small local sub-voxels.
Inspired by ~\citep{qi2017pointnet++}, 
we utilize the inverse distance weighted strategy to interpolate the features of the $t^{th}$ sub-voxel by considering its three nearest neighbors from $\mathcal{Y}_k$, where $t\in \{1, \dots, n_x\times n_y\times n_z\}$ indicating the index of each sub-voxel and we denote its corresponding sub-voxel center as $v_t\in \mathbb{R}^3$. 
Then we can generate the features of the $t^{th}$ sub-voxel as 
\begin{align}
	h^{(v)}_t=\frac{\sum_{i\in \mathcal{G}_t}\left( w_i\cdot h_i\right)}{\sum_{i\in \mathcal{G}_t} w_i}, 
	~~w_{i}=(||a_{i} - v_t||)^{-1}, 
\end{align} 
where $\left[h_i, a_{i}\right] \in \mathcal{Y}_k$, $\mathcal{G}_t$ is the index set indicating the three nearest neighbors (\emph{i.e.}, $(|\mathcal{G}_t|=3)$) of $v_t$ in neighboring set $\mathcal{Y}_k$.
The results $h^{(v)}_t$ encode the local features of the specific $t^{th}$ local sub-voxel in this local cubic.

There are also two other alternative strategies to aggregate the features of local sub-voxels by simply averaging the features within each sub-voxel or by randomly choosing one point within each sub-voxel. Both of them generate lots of empty features in the empty sub-voxels, which may degrade the performance. In contrast, our interpolation based strategy can generate more effective features even on empty local voxels.

Those features in different local sub-voxels may represent very different local features. 
Hence, instead of encoding the local features with a shared-parameter MLP as in~\citep{qi2017pointnet++}, we propose to encode different local sub-voxels with separate local kernel weights for capturing position-sensitive features as 
\begin{align}\label{eq:pos_encoding}
	U_t &= \text{Concat}\left(\left\{a_i-v_t\right\}_{i\in \mathcal{G}_t},~ h^{(v)}_{t} \right)\times W_t, 
\end{align}
where $\left\{a_i-v_t\right\}_{i\in \mathcal{G}_t}\in \mathbb{R}^{(3\times 3=9)}$ indicates the relative positions of its three nearest neighbors, $\text{Concat}(\cdot)$ is the concatenation operation to fuse the relative position and features. 
$W_t \in \mathbb{R}^{(9+C_{\text{in}}) \times C_{\text{mid}}}$  is the learnable kernel weights for encoding the specific features of $t^{th}$ local sub-voxel with feature channel $C_{\text{mid}}$, and different positions have different learnable kernel weights for encoding position-sensitive local features.

Finally, we directly sort the local sub-voxel features $U_t$ according to their spatial order along each of 3D axis, and their features are sequentially concatenated to generate the final local vector representation as
\begin{align}
	\mathcal{U}=\text{MLP}\left(\text{Concat} \left(U_1, U_2, \dots, U_{n_x\times n_y\times  n_z}\right)\right),
\end{align}
where $\mathcal{U} \in \mathbb{R}^{C_{\text{out}}}$. The inner sequential concatenation encodes the structure-preserved local features by simply assigning the features of different locations to their corresponding feature channels, which naturally preserves the spatial structures of local features in the neighboring space centered at $q_k$, 
This local vector representation would be finally processed with several MLPs to encode the local features to $C_{\text{out}}$ feature channels for the follow-up processing.

It is worth noting that our VectorPool aggregation module can also be combined with channel reduction technique as in~\citep{sun2018fishnet} to further reduce the computation/memory resources by summarizing the input feature channels before conducting VectorPool aggregation, and we provide the detailed ablation experiments in Sec.~\ref{sec:ab_study} and Table~\ref{tab:exp_vp_cr}.

Compared with set abstraction, our VectorPool aggregation can greatly reduce the needed computations and memory resources by adopting channel summation and utilizing the proposed local vector representation before MLPs. 
Moreover, instead of conducting max-pooling on local point-wise features as in set abstraction, 
our proposed local vector representation can encode the position-sensitive features with different feature channels, to provide more effective representation for local feature learning.  

\noindent
\textbf{VectorPool Aggregation on PV-RCNN++.}~
Our proposed VectorPool aggregation is integrated in PV-RCNN++ detection framework, to replace set abstraction in both voxel set abstraction module and RoI-grid pooling module. 
Thanks to our VectorPool aggregation operation, the experiments demonstrate that our PV-RCNN++ not only consumes much less memory and computation resources than PV-RCNN framework, but also achieves better 3D detection performance.

\begin{table*}
	\begin{center}
		\scalebox{0.82}{
			\setlength\tabcolsep{2pt}
			\begin{tabular}{l||cc|cc|cc|cc|cc|ccl}
				\hline
				\multirow{2}{*}{Method} & 
			
				\multicolumn{2}{c|}{~~ Veh. (LEVEL 1)  ~~} & \multicolumn{2}{c|}{~Veh. (LEVEL 2)~} & 
				\multicolumn{2}{c|}{~~Ped. (LEVEL 1) ~~} & \multicolumn{2}{c|}{~Ped. (LEVEL 2)~} & 
				\multicolumn{2}{c|}{~Cyc. (LEVEL 1)~} & \multicolumn{2}{c}{Cyc. (LEVEL 2)}\\
				&mAP & mAPH & mAP & \underline{mAPH} & mAP & mAPH & mAP & \underline{mAPH} & mAP & mAPH & mAP & \underline{mAPH}\\
				\hline 
				$^\dagger$SECOND~\citep{yan2018second} & 72.27 & 71.69 & 63.85 & 63.33 & 68.70 & 58.18 & 60.72 & 51.31 & 	60.62 & 59.28 & 58.34 & 57.05 \\ 
				StarNet \citep{ngiam2019starnet} & 53.70 & - & - & - & 66.80 & - & - & - & - & - & - & - \\ 
				$^*$PointPillar \citep{lang2018pointpillars} &  56.62 & - & - & - & 59.25 & - & - & - & - & - & - & -\\ 
				MVF \citep{zhou2020end} & 62.93 & - & - & - & 65.33 & - & - & - & - & - & - & - \\ 
				Pillar-based \citep{wang2020pillar} &  69.80 & - & - & - & 72.51 & - & - & - & - & - & - & - \\
				$^\dagger$Part-A2-Net~\citep{shi2020part}  &  77.05 & 76.51 & 68.47 & 67.97 & 75.24 & 66.87 & 66.18 & 58.62 & 	68.60 & 67.36 & 66.13 & 64.93 \\
				LiDAR R-CNN~\citep{li2021lidar} & 76.0 & 75.5 & 68.3 & 67.9 & 71.2 & 58.7 & 63.1 & 51.7 & 68.6 & 66.9 & 66.1 & 64.4 \\
				$^\dagger$Voxel R-CNN~\citep{deng2021voxel} & 76.13 & 75.66	& 68.18 & 67.74 & 78.20 & 71.98 & 	69.29 & 63.59 & 70.75 & 69.68 & 68.25 &67.21\\
				CenterPoint~\citep{yin2021center} & 76.7 & 76.2 & 68.8 & 68.3 & 79.0 & 72.9 & 71.0 & 65.3 & - & - & - & -\\
				$^\ddagger$CenterPoint~\citep{yin2021center} & - & - & - & 67.9 & -  & - & - & 65.6	& - & - & - & 68.6\\
				RSN~\citep{sun2021rsn} & 75.1 & 74.6 & 66.0 & 65.5 &  77.8 & 72.7 & 68.3 & 63.7 & - & - & - & - \\
				VoTr-TSD~\citep{mao2021voxel} & 74.95 & 74.25 & 65.91 & 65.29& - & - & - & -& - & - & - & -\\
				CT3D~\citep{sheng2021} & 76.30 & - & 69.05 & - & - & - & - & -& - & - & - & -  \\
				PV-RCNN (Ours) & 78.00 & 77.50 &	69.43 & 68.98 & 79.21 & 73.03 & 70.42 & 64.72	& 71.46 & 70.27 & 68.95 & 67.79\\
				PV-RCNN++ (Ours)  & \textbf{79.25} & \textbf{78.78} & 	\textbf{70.61} & \textbf{70.18} & \textbf{81.83} & \textbf{76.28} & \textbf{73.17} & \textbf{68.00} & \textbf{73.72} & \textbf{72.66} & \textbf{71.21} & \textbf{70.19} \\ 
				\hline
				3D-MAN+16f~\citep{yang20213d} & 74.53 & 74.03 & 67.61 & 67.14 & - & - & - & -& - & - & - & - \\
				RSN+3f~\citep{sun2021rsn} & 78.4 & 78.1 & 69.5 & 69.1 & 79.4 & 76.2 &  69.9 & 67.0 & - & - & - & - \\
				PV-RCNN++2f (Ours) & \textbf{80.17} & \textbf{79.70} & \textbf{72.14} & \textbf{71.70} & \textbf{83.48} & \textbf{80.42} & 	\textbf{75.54} & \textbf{72.61} & \textbf{74.63} & \textbf{73.75} & \textbf{72.35} & \textbf{71.50}\\
				\hline
		\end{tabular}}
	\end{center}
    \vspace{-3mm}
	\caption{Performance comparison on the validation set of Waymo Open Dataset. 
		$*$: re-implemented by \citep{zhou2020end}. 
		$\dagger$: re-implemented by ourselves. 
		$\ddagger$: performance reported in the official open-source codebase of \citep{yin2021center}.
		``2f'', ``3f'', ``16f'': the performance is achieved by using multiple point cloud frames. 
	}
	\label{tab:waymo_val}
	\vspace{-2mm}
\end{table*}

\begin{table*}
	\begin{center}
		\scalebox{0.82}{
			\setlength\tabcolsep{2pt}
			\begin{tabular}{l||cc|cc|cc|cc|cc|ccl}
				\hline
				\multirow{2}{*}{Method} & 
			
				\multicolumn{2}{c|}{~~ Veh. (LEVEL 1)  ~~} & \multicolumn{2}{c|}{~Veh. (LEVEL 2)~} & 
				\multicolumn{2}{c|}{~~Ped. (LEVEL 1) ~~} & \multicolumn{2}{c|}{~Ped. (LEVEL 2)~} & 
				\multicolumn{2}{c|}{~Cyc. (LEVEL 1)~} & \multicolumn{2}{c}{Cyc. (LEVEL 2)}\\
				&mAP & mAPH & mAP & \underline{mAPH} & mAP & mAPH & mAP & \underline{mAPH} & mAP & mAPH & mAP & \underline{mAPH}\\
				\hline 
				StarNet \citep{ngiam2019starnet} & 61.5 & 61.0 & 54.9 & 54.5 & 67.8 & 59.9 & 61.1 & 54.0 & - & - & - & - \\ 
				$^*$PointPillar \citep{lang2018pointpillars} &  63.3 & 62.8 & 55.6 & 55.1 & 62.1 & 50.2 & 55.9 & 45.1 & - & - & - & -\\ 
				CenterPoint~\citep{yin2021center} & 80.2 & 79.7 & 72.2 & 71.8 & 78.3 & 72.1 & 72.2 & 66.4 & - & - & - & - \\
				$^\ddagger$CenterPoint~\citep{yin2021center} & - & - & - & 71.9 & -  & - & - & 67.0	& - & - & - & \textbf{68.2}\\
				PV-RCNN (Ours) & 80.60 & 80.15 & 72.81 & 72.39 & 78.16 & 72.01 & 71.81 & 66.05 & 71.80 & 70.42 & 69.13 & 67.80\\
				PV-RCNN++ (Ours) & \textbf{81.62} & \textbf{81.20} & \textbf{73.86} & \textbf{73.47} & \textbf{80.41} & \textbf{74.99} & \textbf{74.12} & \textbf{69.00} & \textbf{71.93} & \textbf{70.76} & \textbf{69.28} & \textbf{68.15}\\
				\hline 
				3D-MAN+16f~\citep{yang20213d} & 78.71 & 78.28 & - & - & 69.97 & 65.98 & - & -& - & - & - & - \\
				CenterPoint+2f~\citep{yin2021center} & 81.05 & 80.59 & 73.42 &	72.99 & 80.47 & 77.28 & 74.56 & 71.52 & \textbf{74.60} &	\textbf{73.68} & \textbf{72.17} & \textbf{71.28} \\
				RSN+3f~\citep{sun2021rsn} & 80.7 & 80.3 & 71.9 & 71.6 & 78.9 & 75.6 & 70.7 & 67.8 & - & - & - & - \\
				PV-RCNN++2f (Ours) & \textbf{83.74} & \textbf{83.32}	& \textbf{76.31} & \textbf{75.92} & \textbf{82.60} & \textbf{79.38} & \textbf{76.63} & \textbf{73.55} & 74.44 & 73.43 & 72.06 & 71.09\\
				\hline
		\end{tabular}}
	\end{center}
    \vspace{-3mm}
	\caption{Performance comparison on the test set of Waymo Open Dataset by submitting to the official test evaluation server. 
		$*$: re-implemented by \citep{zhou2020end}. $\ddagger$: performance reported in the official open-source codebase of \citep{yin2021center}.
		``2f'', ``3f'': the performance is achieved by using multiple point cloud frames.
	}
	\label{tab:waymo3d_test}
	\vspace{-3mm}
\end{table*}

\section{Experiments}
In this section, we first introduce our experimental setup and implementation details in Sec.~\ref{sec:exp_setup}. 
Then we present the main results of our PV-RCNN/PV-RCNN++ frameworks and compare with state-of-the-art methods in Sec.~\ref{sec:main_results}. 
Finally, we conduct extensive ablation experiments and analysis to investigate the individual components of our proposed frameworks in Sec.~\ref{sec:ab_study}.

\subsection{Experimental Setup}\label{sec:exp_setup}

\noindent
\textbf{Datasets and Evaluation Metrics\footnote{The datasets generated during and/or analysed during the current study are available on the official websites of Waymo Open Dataset~\citep{Sun_2020_CVPR} and KITTI dataset~\citep{Geiger2012CVPR}.}.}~
We evaluate our methods on the Waymo Open Dataset~\citep{Sun_2020_CVPR}, which is currently the largest dataset with LiDAR point clouds for 3D object detection of autonomous driving scenarios. There are totally $798$ training sequences with around $160$k LiDAR samples, $202$ validation sequences with $40$k LiDAR samples and $150$ testing sequences with $30$k LiDAR samples.

The evaluation metrics are calculated by the official evaluation tools, where the mean average precision (mAP) and the mAP weighted by heading (mAPH) are used for evaluation. The 3D IoU threshold is set as $0.7$ for vehicle detection and $0.5$ for pedestrian/cyclist detection.
The comparison is conducted in two difficulty levels, where the LEVEL 1 denotes the ground-truth objects with at least 5 inside points while the LEVEL 2 denotes the ground-truth objects with at least 1 inside points. 
As utilized by the official Waymo evaluation server, the mAPH of LEVEL 2 difficulty is the most important evaluate metric for all experiments.

\noindent
\textbf{Network Architecture.}~
For the PV-RCNN framework, the 3D voxel CNN has four levels (see Fig.~\ref{fig:framework_v1}) with feature dimensions $16$, $32$, $64$, $64$, respectively. Their two neighboring radii $r_k$ of each level in the voxel set abstraction module are set as $(0.4\text{m}, 0.8\text{m})$, $(0.8\text{m},1.2\text{m})$, $(1.2\text{m}, 2.4\text{m})$, $(2.4\text{m}, 4.8\text{m})$, and 
the neighborhood radii of set abstraction for raw points are $(0.4\text{m}, 0.8\text{m})$. 
For the proposed RoI-grid pooling operation, we uniformly sample $6\times6\times6$ grid points in each 3D proposal and the two neighboring radii $\tilde{r}$ of each grid point are $(0.8\text{m}, 1.6\text{m})$. 

For the PV-RCNN++ framework, we set the maximum extended radius $r^{(s)}=1.6m$ for proposal-centric filtering, and each scene is split into 6 sectors for parallel keypoint sampling. 
Two VectorPool aggregation operations are adopted to the $4\times$ and $8\times$ feature volumes of voxel set abstraction module with the half length $\delta=(1.2m, 2.4m)$ and $\delta=(2.4m, 4.8m)$ respectively, and both of them have local voxels $n_x=n_y=n_z=3$.  
The VectorPool aggregation on raw points is set with $n_x=n_y=n_z=2$. 
For RoI-grid pooling, we adopt the same number of RoI-grid points ($6\times 6\times 6$) as PV-RCNN, and the utilized VectorPool aggregation has local voxels $n_x=n_y=n_z=3$, 
and half length $\delta=(0.8m, 1.6m)$.

\noindent
\textbf{Training and Inference Details.}
Both two frameworks are trained from scratch in an end-to-end manner with ADAM optimizer, learning rate 0.01 and cosine annealing learning rate decay. To train the proposal refinement stage, we randomly sample 128 proposals with 1:1 ratio for positive and negative proposals, where a proposal is considered as a positive sample if it has at least 0.55 3D IoU with the ground-truth boxes, otherwise it is treated as a negative sample. 
Both two frameworks are trained with three losses with equal loss weights (\emph{i.e.},
region proposal loss, keypoint segmentation loss and proposal refinement loss), where the region proposal loss is same as \citep{yin2021center} and the proposal refinement loss is same as \citep{shi2020part}.

During training, we adopt the widely used data augmentation strategies for 3D detection, including random scene flipping, global scaling with a scaling factor sampled from $[0.95, 1.05]$, global rotation around $Z$ axis with an angle sampled from $[-\frac{\pi}{4}, \frac{\pi}{4}]$, and the ground-truth sampling augmentation~\citep{yan2018second} to randomly ''paste'' some new objects from other scenes to current training scene for simulating objects in various environments.
The detection range is set as $[-75.2, 75.2]m$ for $X$ and $Y$ axes, and $[-2, 4]m$ for the $Z$ axis, while the voxel size is set as $(0.1m, 0.1m, 0.15m)$. 
More training details can be found in our open source codebase \url{https://github.com/open-mmlab/OpenPCDet}.

For the inference speed, our PV-RCNN++ framework can achieve state-of-the-art performance with 10 FPS for $150m \times 150m$ detection range on Waymo Open Dataset (three times faster than PV-RCNN), where a single TITAN RTX GPU card is utilized for profiling.

\subsection{Main Results}\label{sec:main_results}
In this section, we demonstrate the main results of our proposed PV-RCNN/PV-RCNN++ frameworks, and make the comparison with state-of-the-art methods on the large-scale Waymo Open Dataset~\citep{Sun_2020_CVPR}. 
By default, we adopt the center-based RPN head as in \citep{yin2021center} to generate 3D proposals in the first stage, and we train a single model in each setting for detecting the objects of all three categories.

\noindent
\textbf{Comparison with State-of-the-Art Methods.}~
As shown in Table~\ref{tab:waymo_val}, 
for the 3D object detection setting of taking a single frame point cloud as input, 
our PV-RCNN++ (\emph{i.e.}, ``PV-RCNN++'') outperforms previous state-of-the-art works~\citep{yin2021center,shi2020part} 
on all three categories with remarkable performance gains (+1.88\% for vehicle, 
+2.40\% for pedestrian and 
+1.59\% for cyclist
in terms of mAPH of LEVEL 2 difficulty).
Moreover, following \citep{sun2021rsn}, by simply concatenating an extra neighboring past frame as input, our PV-RCNN framework can also be evaluated on the multi-frame setting. Table~\ref{tab:waymo_val} (\emph{i.e.}, ``PV-RCNN++2f'') demonstrates that the performance of our PV-RCNN++ framework can be further boosted by using 2 frames, which outperforms previous multi-frame method~\citep{sun2021rsn} with remarkable margins (+2.60\% for vehicle, +5.61\% for pedestrian in terms of mAPH of LEVEL 2). 
 
Meanwhile, we also evaluate our frameworks on the test set by submitting to the official test server of Waymo Open Dataset~\citep{Sun_2020_CVPR}. As shown in Table~\ref{tab:waymo3d_test}, without bells and whistles, in both single-frame and multi-frame settings, our PV-RCNN++ framework consistently outperforms previous state-of-the-art~\citep{yin2021center} significantly in both vehicle and pedestrian categories, where for single-frame setting we achieve a performance gain of 
+1.57\% for vehicle and +2.00\% for pedestrian 
in terms of mAPH of LEVEL 2 difficulty, and for multi-frame setting we achieve a performance gain of +2.93\% for vehicle detection and +2.03\% for pedestrian detection. 
We also achieve comparable performance for the cyclist category on both the single-frame and multi-frame settings.
Note that we do not use any test-time augmentation or model ensemble tricks in the evaluation process. 
The significant improvements on the large-scale Waymo Open dataset manifest the effectiveness of our proposed framework.

\begin{table}
	\begin{center}
		\scalebox{0.85}{
			\begin{tabular}{c|c|c|c|c}
				\hline
				\multirow{2}{*}{Difficulty} &
				\multirow{2}{*}{Method} &
				\multicolumn{1}{c|}{Veh.} & 
				\multicolumn{1}{c|}{Ped.}& 
				\multicolumn{1}{c}{Cyc.}\\
				& & mAPH& mAPH& mAPH\\ 
				\hline
				\multirow{4}{*}{LEVEL 1} & PV-RCNN (anchor) & 76.89 & 65.65 & 66.35 \\
				& PV-RCNN++ (anchor) & \textbf{78.64} & \textbf{69.26} & \textbf{70.86}\\
				\cline{2-5}
				
				& PV-RCNN (center) & 77.50 & 73.03 & 70.27 \\
				
				& PV-RCNN++ (center) & \textbf{78.63} & \textbf{74.62} & \textbf{72.38} \\
				\hline 
				\multirow{4}{*}{LEVEL 2} & PV-RCNN (anchor) & 68.41 & 57.61 & 63.98\\
				& PV-RCNN++ (anchor) & \textbf{69.95} & \textbf{60.94} & \textbf{68.22}\\
				\cline{2-5}
				& PV-RCNN (center) & 68.98 & 64.72 & 67.79\\
				& PV-RCNN++ (center) & \textbf{69.91} & \textbf{66.30} & \textbf{69.62} \\
				\hline 
		\end{tabular}}
	\end{center}
\vspace{-2mm}
	\caption{Performance comparison of PV-RCNN and PV-RCNN++ on the validation set of Waymo Open Dataset. We adopt two settings for both two frameworks by equipping with different RPN heads for proposal generation, which are the anchor-based RPN head as in \citep{shi2020part} and the center-based RPN head as in \citep{yin2021center}. Note that PV-RCNN++ adopts the same backbone network (without residual connection) with PV-RCNN for fair comparison.
	}
	\label{tab:ab_pvrcnn++}
	\vspace{-2mm}
\end{table}

\begin{table}
	\begin{center}
		\scalebox{0.95}{
			\begin{tabular}{c|ccc|c}
				\hline
				Method & Car & Pedestrian & Cyclist & Average \\
				\hline
				PV-RCNN & 81.43 & 43.29 & 63.71 & 62.81 \\
				PV-RCNN++ & {\bf 81.88} & {\bf 47.19} & {\bf 67.33} & {\bf 65.47} \\
				\hline 
		\end{tabular}}
	\end{center}
	\vspace{-2mm}
	\caption{Performance comparison of PV-RCNN and PV-RCNN++ on the test set of KITTI dataset. The results are evaluated by the most important moderate difficulty level of KITTI evaluation metric by submitting to the official KITTI evaluation server.
	}
	\label{tab:ab_pvrcnn++_kitti}
	\vspace{-2mm}
\end{table}

\noindent
\textbf{Comparison of PV-RCNN and PV-RCNN++.}
Table~\ref{tab:ab_pvrcnn++} demonstrates that no matter which type of RPN head is adopted, our PV-RCNN++ framework consistently outperforms previous PV-RCNN framework on all three categories of all difficulty levels. 
Specifically, for the anchor-based setting, PV-RCNN++ surpasses PV-RCNN with a performance gain of +1.54\% for vehicle, +3.33\% for pedestrian and 4.24\% for cyclist in terms of mAPH of LEVEL 2 difficulty. 
By taking the center-based head, PV-RCNN++ also outperforms PV-RCNN with a +0.93\% mAPH gain for vehicle, a +1.58\% mAPH gain for pedestrian and a +1.83\% mAPH gain for cyclist in terms of LEVEL 2 difficulty.

The stable and consistent improvements prove the effectiveness of our proposed sectorized proposal-centric sampling algorithm and VectorPool aggregation module. 
More importantly, our PV-RCNN++ consumes much less calculations and GPU memory than PV-RCNN framework, while also increasing the processing speed from 3.3 FPS to 10 FPS for the 3D detection of $150m \times 150m$ such a large area, which further validates the efficiency and the effectiveness of our PV-RCNN++. 

As shown in Table~\ref{tab:ab_pvrcnn++_kitti}, we also provide the performance comparison of PV-RCNN and PV-RCNN++ on the KITTI dataset~\citep{Geiger2012CVPR}. Compared with Waymo Open Dataset, KITTI dataset adopts different kinds of LiDAR sensor and the scene in KITTI dataset is about four times smaller than the scene in the Waymo Open Dataset. 
Table~\ref{tab:ab_pvrcnn++} shows that PV-RCNN++ outperforms previous PV-RCNN on all three categories of KITTI dataset with remarkable average performance margin, demonstrating its effectiveness on handling different kinds of scenes and different LiDAR sensors.

\subsection{Ablation Study}\label{sec:ab_study}
In this section, we investigate the individual components of our PV-RCNN++ framework with extensive ablation experiments. 
We conduct all experiments on the large-scale Waymo Open Dataset~\citep{Sun_2020_CVPR}.
For efficiently conducting the ablation experiments, we generate a small representative training set by uniformly sampling $20\%$ frames (about $32k$ frames) from the training set\footnote{Reference: https://github.com/open-mmlab/OpenPCDet.}, and all results are evaluated on the full validation set (about $40k$ frames) with the official evaluation tool. All models are trained with 30 epochs and batch size 16 on 8 GPUs. 

We conduct all ablation experiments with the center-based RPN head~\citep{yin2021center} on three categories (vehicle, pedestrian and cyclist) of Waymo Open Dataset~\citep{Sun_2020_CVPR}, and the mAPH of LEVEL 2 difficulty is adopted as the evaluation metric.

\begin{table}[t]
	\begin{center}
		\scalebox{0.92}{
			\setlength\tabcolsep{2pt}
			\begin{tabular}{cc|ccc|c}
				\hline
				{\tabincell{c}{Point Feature \\Extraction}} & {\tabincell{c}{RoI Pooling \\ Module}} & Veh. & Ped.
				& Cyc. & Average \\
				\hline
				UNet-decoder & RoI-aware Pooling & 66.42 & 63.41 & 67.48 & 65.77  \\
				UNet-decoder & RoI-grid Pooling & 67.37 & 63.77 & 67.08 & 66.05 \\ 
				\hline
				VSA & RoI-aware Pooling & 66.15 & 60.70 & 66.07 & 64.31 \\
				VSA & RoI-grid Pooling & 68.62  & 63.74 & 68.26 & 66.87 \\
				\hline 
			\end{tabular}
	}
	\end{center}
	\vspace{-2mm}
	\caption{Effects of voxel set abstraction (VSA) and RoI-grid pooling modules, where the  UNet-decoder and RoI-aware pooling are the same with \citep{shi2020part}. All experiments are based on PV-RCNN++ framework with a center-based RPN head.
	}
	\label{tab:exp_vsa_roigrid}
	\vspace{-2mm}
\end{table}

\begin{table}[t]
	\begin{center}
			\scalebox{0.86}{
				 			\setlength\tabcolsep{2pt}
\begin{tabular}{cc|cc|c|ccc|c}
	\hline
	\multicolumn{4}{c|}{VSA Input Feature} & \multirow{2}{*}{\tabincell{c}{Frame\\Rate}} & \multirow{2}{*}{Veh.} & \multirow{2}{*}{Ped.}
	& \multirow{2}{*}{Cyc.} & \multirow{2}{*}{Average} \\ 
	\cline{1-4}
	$f_i^{(\text{pv}_{1, 2})}$ & $f_i^{(\text{pv}_{3, 4})}$ & $f_i^{(\text{bev})}$ & 
	$f_i^{(\text{raw})}$
	& & & &  \\
	\hline
	& &\checkmark& & 16.3 & 67.01 & 61.23 & 66.54 & 64.93 \\
	& & & \checkmark & 13.2 & 68.02 & 62.73 & 66.80 & 65.85 \\
	&\checkmark& & & 12.3 & 68.13 & 63.24 & 67.37 &  66.25 \\
	\hline
	&\checkmark&\checkmark& & 11.9 & 68.15 & 63.35 & 67.73 & 66.41 \\  
	& \checkmark&& \checkmark& 10.3 & 68.46 & 63.59 & 67.58 & 66.54 \\
	&\checkmark&\checkmark& \checkmark & 10.0 & 68.62 & 63.74 & 68.26 & 66.87  \\  
	\checkmark&\checkmark&\checkmark&\checkmark& \textit{7.6} & \textit{68.55} & \textit{64.48} & \textit{67.94} & \textit{66.99}\\
	\hline
\end{tabular}
		}
	\end{center}
	\vspace{-2mm}
	\caption{Effects of different feature components for voxel set abstraction. ``Frame Rate'' indicates frames per seconds in terms of testing speed. All experiments are conducted on PV-RCNN++ framework with a center-based RPN head. Note that the default setting of PV-RCNN++ does not use the voxel features $f_i^{(\text{pv}_{1, 2})}$ by considering its negligible gain and higher latency.
	}
	\label{tab:exp_vsa}
	\vspace{-2mm}
\end{table}

\noindent
\textbf{Effects of Voxel-to-Keypoint Scene Encoding.}
In Sec.~\ref{sec:vsa}, we propose the voxel-to-keypoint scene encoding strategy to encode the global scene features to a small set of keypoints, which serves as a bridge between the backbone network and the proposal refinement network.
As shown in the $2^{nd}$ and $4^{th}$ rows of Table~\ref{tab:exp_vsa_roigrid}, 
our proposed voxel-to-keypoint scene encoding strategy achieves 
better performance than the UNet-based decoder while summarizing the scene features to much less point-wise features than the UNet-based decoder.
For instance, our voxel set abstraction module encodes the whole scene to around $4k$ keypoints for feeding into the RoI-grid pooling module, while the UNet-based decoder network needs to summarize the scene features to around $80k$ point-wise features in most cases, which validates the effectiveness of our proposed voxel-to-keypoint scene encoding strategy. 
We consider that it might benefit from the fact that the keypoint features are aggregated from multi-scale feature volumes and raw point clouds with large receptive fields,  
while also keeping the accurate point locations. 
Besides that, we should also note that the feature dimension of UNet-based decoder is generally smaller than the feature dimensions of our keypoints since the UNet-based decoder is limited to its large memory consumption on large-scale point clouds, which may degrade its performance.

We also notice that our voxel set abstraction module achieves worse performance (the $1^{st}$ and $3^{rd}$ rows of Table~\ref{tab:exp_vsa_roigrid}) than the UNet-decoder when it is combined with RoI-aware pooling~\citep{shi2020part}. This is to be expected since RoI-aware pooling module will generate lots of empty voxels in each proposal by taking only $4k$ keypoints, which may degrade the performance. 
In contrast, our voxel set abstraction module can be ideally combined with our RoI-grid pooling module and they can benefit each other by taking a small number of keypoints as the intermediate connection.

\noindent
\textbf{Effects of Different Features for Voxel Set Abstraction.}
The voxel set abstraction module incorporates multiple feature components (see Sec.~\ref{sec:vsa} ), and their effects are explored in Table~\ref{tab:exp_vsa}. We can summarize the observations as follows: 
$(i)$ The performance drops a lot if we only aggregate features from high level bird-view semantic features ($f_i^{(\text{bev})}$) or accurate point locations ($f_i^{(\text{raw})}$), since neither 2D-semantic-only nor point-only are enough for the proposal refinement. 
$(ii)$ As shown in 6$^{th}$ row of Table~\ref{tab:exp_vsa}, $f_i^{(\text{pv}_3)}$ and $f_i^{(\text{pv}_4)}$ contain both 3D structure information and high level semantic features, which can improve the performance a lot by combining with the bird-view semantic features $f_i^{(\text{bev})}$ and the raw point locations $f_i^{(\text{raw})}$. 
$(iii)$ The shallow semantic features $f_i^{(\text{pv}_1)}$ and $f_i^{(\text{pv}_2)}$ can slightly improve the performance but also greatly increase the training cost. Hence, the proposed PV-RCNN++ framework does not use such shallow semantic features.

\begin{table}
	\begin{center}
		\scalebox{0.9}{
			\begin{tabular}{c|cccc}
				\hline
				\multirow{1}{*}{Use PKW} & Vehicle & Pedestrian & Cyclist & Average\\ 
				\hline
				\xmark & 68.48 & 63.90 & 67.62 & 66.66 \\
				\checkmark & 68.62 & 63.74 & 68.26 & 66.87 \\
				\hline
			\end{tabular}
 		}
	\end{center}
	\caption{Effects of Predicted Keypoint Weighting module.  All experiments are conducted on our PV-RCNN++ framework with a center-based RPN head.}
	\label{tab:exp_pkw}
	\vspace{-2mm}
\end{table}

\noindent
\textbf{Effects of Predicted Keypoint Weighting.} 
The predicted keypoint weighting is proposed in Sec.~\ref{sec:vsa} to re-weight the point-wise features of keypoints with extra keypoint segmentation supervision. 
As shown in Table~\ref{tab:exp_pkw}, the experiments 
show that the performance slightly drops after removing this module, which demonstrates that the predicted keypoint weighting enables better multi-scale feature aggregation by focusing more on the foreground keypoints, since they are more important for the succeeding proposal refinement network. 
Although this module only leads small additional cost to our frameworks, we should also notice that it is optional for our frameworks by considering its limited gains.

\begin{table*}
	\small 
	\begin{center}
			\scalebox{0.9}{
				\begin{tabular}{l|c|cccc|ccccc|c}
					\hline
					\multirow{2}{*}{\tabincell{c}{Keypoint Sampling \\ Algorithm}} & \multirow{2}{*}{\tabincell{c}{Running \\ Time}} & \multicolumn{4}{c|}{Performance of L2 mAPH}
					&\multicolumn{5}{c|}{Coverage Rate (CR) under Different Radii}& \multirow{2}{*}{\tabincell{c}{Mean\\ CR}}\\ 
					& & Veh. & Ped.
					& Cyc. & Average & 0.1m & 0.2m & 0.3m & 0.4m & 0.5m \\
					\hline
					FPS & 133ms & 67.70 & 63.18 & 66.74 & 65.87 & - & - & -& - & - & -   \\
					PC-Filter+FPS & 27ms & 68.78 & 64.09 & 68.11 & 66.99 & 43.22 & 82.78 & 97.97 & 99.95 & 99.99 & 84.78 \\
					\hline
					PC-Filter+Random Sampling & $<$1ms & 65.10 & 60.59 & 65.57& 63.75 &51.44 & 77.06 & 87.32 & 92.16 & 94.74 & 80.54\\ 
					PC-Filter+Voxelized-FPS-Voxel & 17ms & 68.01 & 63.21 & 67.36 & 66.19 & 6.52 & 52.00 & 89.93 & 98.89 & 99.92& 69.45 \\
					PC-Filter+Voxelized-FPS-Point & 17ms & 68.37 & 63.45 & 67.30 & 66.37 & 34.14 & 75.84 & 96.28 & 99.77 & 99.99 & 81.20 \\ 
					PC-Filter+RandomParallel-FPS & 2ms & 68.05 & 63.20 & 66.79 & 66.01 & 38.12 & 64.47 & 81.72	& 91.82 & 96.90	& 74.61 \\ 
					PC-Filter+Sectorized-FPS  & 9ms & 68.62 & 63.74 & 68.26 & 66.87 & 47.63 & 82.38 & 95.20 &	98.87 & 99.72	& 84.76 \\
					\hline
		\end{tabular}}
	\end{center}
	\vspace{-2mm}
	\caption{Effects of different keypoint sampling algorithms. The running time is the average running time of keypoint sampling process on the validation set of the Waymo Open Dataset. The coverage rate is calculated by averaging the coverage rate of each scene on the validation set of the Waymo Open Dataset. ``FPS'' indicates the farthest point sampling and ``PC-Filter'' indicates our proposal-centric filtering strategy. All experiments are conducted by adopting different keypoint sampling algorithms to our PV-RCNN++ framework with a center-based RPN head.}
	\label{tab:exp_spc}
	\vspace{-2mm}
\end{table*}

\begin{figure*}
	\begin{center}
		\includegraphics[width=1.0\linewidth]{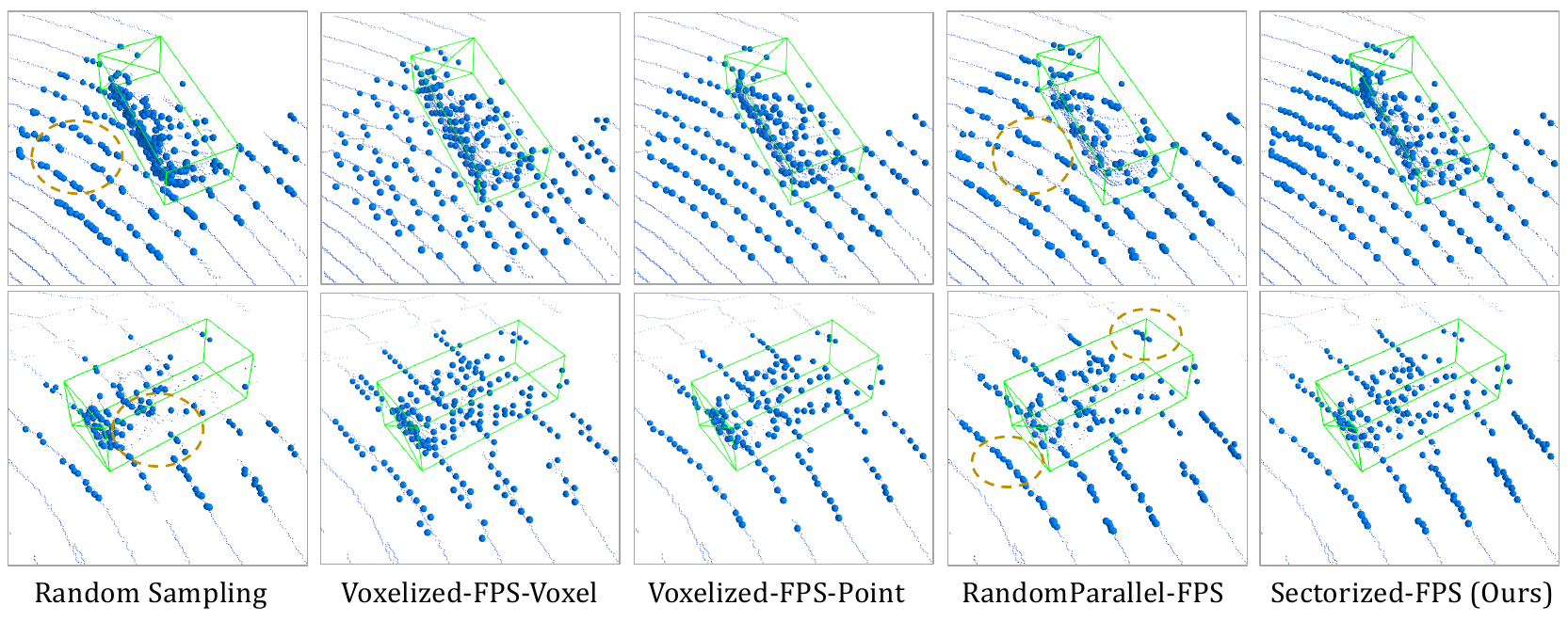}
	\end{center}
	\vspace{-3.9mm}
	\caption{Illustration of the keypoint distributions from different keypoint sampling strategies. Some dashed circles are utilized to highlight the missing parts and the clustered keypoints after using these keypoint sampling strategies.
		We find that our Sectorized-FPS generates better uniformly distributed keypoints that cover more input points to better encode the scene features for proposal refinement, while other strategies may miss some important regions or generate some clustered keypoints.
	}
	\label{fig:spc}
\end{figure*}

\noindent
\textbf{Effects of RoI-grid Pooling Module.}
RoI-grid pooling module is proposed in Sec.~\ref{sec:roi_grid_pool} for aggregating RoI features from the very sparse keypoints.
Here we investigate the effects of RoI-grid pooling module by replacing it with the RoI-aware pooling \citep{shi2020part} and keeping other modules consistent. 
As shown in  the $3^{rd}$ and $4^{th}$ rows Table~\ref{tab:exp_vsa_roigrid}, 
the experiments 
show that the performance drops significantly when replacing RoI-grid pooling. It validates that our proposed RoI-grid pooling module can aggregate much richer contextual information to generate more discriminative RoI features.

Compared with the previous RoI-aware pooling module~\citep{shi2020part}, our proposed RoI-grid pooling module can generate denser grid-wise feature representation by supporting different overlapped ball areas among different grid points, while RoI-aware pooling module may generate lots of zeros due to the sparse inside points of RoIs. 
That means our proposed RoI-grid pooling module is especially effective for aggregating local features from the very sparse point-wise features, such as in our PV-RCNN framework to aggregate features from a very small number of keypoints.

\noindent
\textbf{Effects of Proposal-Centric Filtering.}~ 
In the $1^{st}$ and $2^{nd}$ rows of Table~\ref{tab:exp_spc}, we investigate the effectiveness of our proposal-centric keypoint filtering (see Sec.~\ref{sec:spc}), 
where we find that compared with the strong baseline PV-RCNN++  framework equipped with vanilla farthest point sampling, our proposal-centric keypoint filtering further improves the average detection performance by $1.12$ mAPH in LEVEL 2 difficulty (65.87\% vs. 66.99\%).
It validates our argument that our proposed proposal-centric keypoint sampling strategy can generate more representative keypoints by concentrating the small number of keypoints to the more informative neighboring regions of proposals. Moreover, improved by our proposal-centric keypoint filtering, our keypoint sampling algorithm is about five times (133ms vs. 27ms) faster than the vanilla farthest point sampling algorithm by reducing the number of candidate keypoints.

\begin{table*}
	\begin{center}
		\scalebox{0.78}{
			\begin{tabular}{c|ccc|cccccc|cc|c|c}
				\hline
				\multirow{2}{*}{\tabincell{c}{Setting}} & \multirow{2}{*}{\tabincell{c}{Keypoint \\ Sampling}} & \multirow{2}{*}{\tabincell{c}{Point Feature \\ Extraction}} & \multirow{2}{*}{\tabincell{c}{RoI Feature \\ Aggregation}} & \multicolumn{2}{c}{Vehicle} & \multicolumn{2}{c}{Pedestrian} & \multicolumn{2}{c|}{Cyclist} & \multicolumn{2}{c|}{Average} & \multirow{2}{*}{GFLOPS} & \multirow{2}{*}{\tabincell{c}{Memory\\ (MB)}} \\
				& & & & L1 & L2 & L1 & L2 & L1 & L2 & L1 & L2 & &\\
				\hline
				PV-RCNN & FPS & VSA (SA) &  RoI-grid pool (SA) & 75.43 & 67.54 & 69.40 & 61.62 & 68.98 & 66.57 & 71.27 & 65.24 &  -0 & 10617 \\
				\hline
				& SPC-FPS & VSA (SA) & RoI-grid pool (SA) & 76.44 & 68.03 & 70.52 & 62.43 & 69.84 & 67.29 & 72.27 & 65.92   & -0 & 10617\\
				& SPC-FPS & VSA (VP) & RoI-grid pool (SA) & 77.41 & 68.73 & 70.98 & 63.30 & 69.63 & 67.19 & 72.67 & 66.41 & -1.712 & 9453\\
				& SPC-FPS & VSA (SA) & RoI-grid pool (VP) & 77.12 & 68.43 & 70.82 & 63.15 & 70.11 & 67.66 & 72.68 & 66.41 & -2.967 & 8565\\ 
				PV-RCNN++ & SPC-FPS &  VSA (VP) & RoI-grid pool (VP) & 77.32 & 68.62 & 71.36 & 63.74 & 70.71 & 68.26 & 73.13 & 66.87 & -4.679 &7583  \\ 
				\hline
		\end{tabular}}
	\end{center}
	\vspace{-1mm}
	\caption{Effects of different components in our proposed PV-RCNN++ frameworks. All models are trained with $20\%$ frames from the training set and are evaluated on the full validation set of the Waymo Open Dataset, and the evaluation metric is the mAPH in terms of LEVEL\_1 (L1) and LEVEL\_2 (L2) difficulties as used in ~\citep{Sun_2020_CVPR}. ``FPS'' denotes farthest point sampling, ``SPC-FPS'' denotes our proposed sectorized proposal-centric keypoint sampling strategy, ``VSA'' denotes the voxel set abstraction module, ``SA'' denotes the set abstraction operation and ``VP'' denotes our proposed VectorPool aggregation. All models adopt the center-based RPN head for proposal generation.}
	\label{tab:exp_improvements}
	\vspace{-2mm}
\end{table*}

\noindent
\textbf{Effects of Sectorized Keypoint Sampling.} 
To investigate the effects of sectorized farthest point sampling (Sec.~\ref{sec:spc}), 
we compare it with four alternative strategies for accelerating the keypoint sampling process: 
$(i)$ Random Sampling: the keypoints are randomly chosen from raw points. 
$(ii)$ Voxelized-FPS-Voxel: the raw points are firstly voxelized to reduce the number of points (\emph{i.e.,} voxels), then farthest point sampling is applied to sample keypoints from voxels by taking the voxel centers. 
$(iii)$ Voxelized-FPS-Point: unlike Voxelized-FPS-Voxel, here a raw point is randomly selected within the selected voxels as the keypoints.  
$(iv)$ RandomParallel-FPS: the raw points are randomly split into several groups, then farthest point sampling is utilized to these groups in parallel for faster keypoint sampling.

As shown in Table~\ref{tab:exp_spc}, compared with the vanilla farthest point sampling ($2^{nd}$ row) algorithm, the detection performances of all four alternative strategies drop a lot. In contrast, the performance of our proposed sectorized farthest point sampling algorithm is on par with the vanilla farthest point sampling (66.99\% vs. 66.87\%) while being three times (27ms vs. 9ms) faster than the vanilla farthest point sampling algorithm.

\noindent
\textbf{Analysis of the Coverage Rate of Keypoints.}
We argue that the uniformly distributed keypoints are important for the proposal refinement, where a better keypoint distribution should cover more input points. 
Hence, to evaluate the quality of different keypoint sampling strategies,
 we propose an evaluation metric, \textit{coverage rate}, 
which is defined as the ratio of input points that are within the coverage region of any keypoints. 
Specifically, for a set of input points $\mathcal{P}=\{p_i\}_{i=1}^m$ and a set of sampled keypoints $\mathcal{K}=\{p'_j\}_{j=1}^n$, the coverage rate $\mathbf{C}$ can be formulated as: 
\begin{align}\label{eq:coverage_rate}
	\mathbf{C} = \frac{1}{m}\cdot{\sum_{i=1}^{m} \min (1.0,~~~ \sum_{j=1}^{n}\mathds{1}({||p_i - p'_j|| < R_c}))},
\end{align}
where $R_c$ is a scalar that denotes the coverage radius of each keypoint, 
and $\mathds{1}\left(\cdot\right)\in\{0, 1\}$ is the indicator function to indicate whether $p_i$ is covered by $p'_j$.

As shown in Table~\ref{tab:exp_spc}, we evaluate the coverage rate of different keypoint sampling algorithms in terms of multiple coverage radii. Our sectorized farthest point sampling achieves similar average coverage rate (84.76\%) with the vanilla farthest point sampling (84.78\%), which is much better than other sampling algorithms.
The higher average coverage rate demonstrates that our proposed sectorized farthest point sampling can sample more uniformly distributed keypoints to better cover the input points, which is consistent with the qualitative results of different sampling strategies as in Fig.~\ref{fig:spc}.

In short, our sectorized farthest point sampling can generate uniformly distributed keypoints to better cover the input points,  by splitting raw points into different groups based on the fact of radial distribution of LiDAR points. 
Although there may still exist a very small number of clustered keypoints in the margins of different sectors, the experiments show that they have negligible effect on the performance. 
We consider the reason may be that the clustered keypoints are mostly in the regions around the scene centers, where the objects are generally easier to detect since the raw points around scene centers are much denser than distant regions.

\noindent
\textbf{Effects of VectorPool Aggregation.}~
In Sec.~\ref{sec:vectorpool}, to tackle the resource-consuming problem of set abstraction, we propose VectorPool aggregation module to effectively and efficiently summarize the structure-preserved local features from point clouds. 
As shown in Table~\ref{tab:exp_improvements}, by adopting VectorPool aggregation in both voxel set abstraction module and RoI-grid pooling module, PV-RCNN++ framework consumes much less computations (\emph{i.e.}, a reduction of 4.679 GFLOPS) and GPU memory (from 10.62GB to 7.58GB) than original PV-RCNN framework, while the performance is also consistently increased from 65.92\% to 66.87\% in terms of average mAPH (LEVEL 2) of three categories. Note that the batch size is only set as 2 in all of our settings and the reduction of memory consumption / calculations can be more significant with larger batch size. 

The significant reduction of resource consumption demonstrates the effectiveness of our VectorPool aggregation for feature learning from large-scale point clouds, which makes our PV-RCNN++ framework a more practical 3D detector to be used on resource-limited devices. Moreover, PV-RCNN++ framework also benefits from the structure-preserved spatial features from our VectorPool aggregation, which is critical for the following fine-grained proposal refinement.

We further analyze the effects of VectorPool aggregation by removing channel reduction~\citep{sun2018fishnet} in our VectorPool aggregation. 
As shown in Table~\ref{tab:exp_vp_cr}, our VectorPool aggregation is effective in reducing memory consumption no matter whether channel reduction is incorporated (by comparing the $1^{st}$ / $3^{rd}$ rows or the $2^{nd}$ / $4^{th}$ rows), since the model activations in our VectorPool aggregation modules consume much less memory than those in set abstraction, by adopting a single local vector representation before multi-layer perceptron networks.  
Meanwhile, Table~\ref{tab:exp_vp_cr} also demonstrates that our VectorPool aggregation can achieve better performance than set abstraction~\citep{qi2017pointnet++} in both two cases (with or without channel reduction).
Meanwhile, we also notice that VectorPool aggregation slightly improves the number of parameters compared with previous set abstraction module (\emph{e.g.}, from 13.05M to 14.11M for the setting with channel reduction), which is generally negligible given the fact that VectorPool aggregation consumes smaller GPU memory.  

\begin{table}
	\begin{center}
		\scalebox{0.9}{
			\setlength\tabcolsep{2pt}
			\begin{tabular}{c|c|ccc|c|c|c|c}
				\hline
				{\tabincell{c}{Stra-\\tegy}} & {\tabincell{c}{CR}} & Veh.  & Ped. & Cyc. & Average & GFLOPS & \tabincell{c}{Mem.\\ (MB)} & \tabincell{c}{\#Param.}\\
				\hline
				SA & \xmark & 68.03 & 62.43 & 67.29 & 65.92  & -0 & 10617 & 13.07M\\
				SA &\checkmark &  68.43	& 62.06 & 66.96& 65.81 & -3.467 & 9795 & 13.05M \\ 
				VP &\xmark & 68.82	& 64.06	& 67.96 & 66.95 & -1.988 & 8549 & 14.32M\\
				VP & \checkmark & 68.62 & 63.74 & 68.26 & 66.87 & -4.679 &7583 & 14.11M  \\ 
				\hline
		\end{tabular}}
	\end{center}
	\caption{Effects of VectorPool aggregation with and without channel reduction~\citep{sun2018fishnet}.  
		``SA'' denotes set abstraction, ``VP'' denotes VectorPool aggregation module and ``CR'' denotes channel reduction. 
		``\#Param.'' indicates the number of parameters of the model.}
		All experiments are based on our PV-RCNN++ framework with a center-based RPN head for proposal generation, and only the local feature extraction modules are changed during the ablation experiments.
	\label{tab:exp_vp_cr}
	\vspace{-3mm}
\end{table}

\begin{table}
	\begin{center}
		\scalebox{0.9}{
			\begin{tabular}{c|ccc|c}
				\hline
				\tabincell{c}{Aggregation \\ of Sub-Voxels} &  Vehicle  & Pedestrian & Cyclist & Average \\
				\hline
				Average Pooling & 68.35 & 62.33 & 67.50 & 66.06\\ 
				Random Selection & 68.36 & 62.82 & 67.68 & 66.29 \\ 
				Interpolation & 68.62 & 63.74 & 68.26 & 66.87 \\
				\hline
		\end{tabular}}
	\end{center}
	\caption{Effects of the feature aggregation strategies to generate the local sub-voxel features of VectorPool aggregation. All experiments are based on our PV-RCNN++ framework with a center-based RPN head for proposal generation.}
	\label{tab:vp_local_pooling}
	\vspace{-3mm}
\end{table}

\noindent
\textbf{Effects of Different Feature Aggregation Strategies for Local Sub-Voxels.}~
As mentioned in Sec.~\ref{sec:vectorpool}, in addition to our adopted interpolation-based method, there are two alternative strategies (average pooling and random selection) for aggregating features of local sub-voxels. 
Table~\ref{tab:vp_local_pooling} demonstrates that 
our interpolation based feature aggregation achieves much better performance than the other two strategies, especially for the small objects like pedestrian and cyclist.   
We consider that our strategy can generate more effective features by interpolating from three nearest neighbors (even beyond the sub-voxel), while both of the other two methods might generate lots of zero features on the empty sub-voxels, which may degrade the final performance.

\begin{table}[t]
	\begin{center}
		\scalebox{0.9}{
			\setlength\tabcolsep{3pt}
			\begin{tabular}{cc|ccc|c}
				\hline
				{\tabincell{c}{Kernel \\ Weights}} & {\tabincell{c}{Number of \\ Dense Voxels}} & Vehicle  & Pedestrian & Cyclist & Average\\
				\hline
				Share & $3\times 3 \times 3$ & 68.17 & 63.28 & 67.36 & 66.27 \\
				Separate & $3\times 3 \times 3$ & 68.62 & 63.74 & 68.26 & 66.87 \\
				\hline 
				Separate & $2\times 2 \times 2$ & 68.21 & 62.88 & 67.44 & 66.18\\ 
				Separate & $3\times 3 \times 3$ & 68.62 & 63.74 & 68.26 & 66.87 \\ 
				Separate & $4\times 4 \times 4$ & 68.74 & 63.99 & 67.98 & 66.90 \\ 
				\hline
		\end{tabular}}
	\end{center}
	\caption{Effects of separate local kernel weights and the number of dense voxels in our proposed VectorPool aggregation module. All experiments are based on our PV-RCNN++ framework with a center-based head for proposal generation.}
	\label{tab:separate_kernel}
\end{table}

\noindent
\textbf{Effects of Separate Local Kernel Weights in VectorPool Aggregation.}~
We adopt separate local kernel weights (see Eq.~\eqref{eq:pos_encoding}) on different local sub-voxels to generate position-sensitive features.
The $1^{st}$ and $2^{nd}$ rows of Table~\ref{tab:separate_kernel} show that the performance drops a bit if we remove the separate local kernel weights and adopt shared kernel weights for relative position encoding. It validates that the separate local kernel weights are better than previous shared-parameter MLP for local feature encoding, and it is important in our proposed VectorPool aggregation module.

\noindent
\textbf{Effects of Dense Voxel Numbers in VectorPool Aggregation.}~
Table~\ref{tab:separate_kernel} investigates the number of dense voxels $n_x\times n_y \times n_z$ in VectorPool aggregation for voxel set abstraction module and RoI-grid pooling module, where we can see that VectorPool aggregation with $3\times 3\times 3$ and $4 \times 4 \times 4$ achieve similar performance while the performance of $2\times 2\times 2$ setting drops a lot. 
We consider that our interpolation-based VectorPool aggregation can generate effective voxel-wise features even with large dense voxels, hence the setting with $4\times 4 \times 4$ achieves slightly better performance than the setting with $3\times 3\times 3$. However, since the setting with $4 \times 4 \times 4$ greatly improves the calculations and memory consumption, we finally choose the setting of $3\times 3\times 3$ dense voxel representation in both voxel set abstraction module (except the raw point layer) and RoI-grid pooling module of our PV-RCNN++ framework. 

\begin{table}
	\begin{center}
		\scalebox{0.9}{
			\begin{tabular}{c|ccc|c}
				\hline
				\tabincell{c}{Number of \\ Keypoints} &  Vehicle  & Pedestrian & Cyclist & Average \\
				\hline
				8192 & 68.85 & 64.11 & 67.88 & 66.95 \\
				4096 & 68.62 & 63.74 & 68.26 & 66.87 \\
				2048 & 67.99 & 62.14 & 67.41 & 65.85 \\
				1024 & 66.67 & 59.21 & 65.07 & 63.65  \\
				\hline
			\end{tabular}
		}
	\end{center}
	\caption{Effects of the number of keypoints for encoding the global scene. All experiments are based on our PV-RCNN++ framework with a center-based head for proposal generation.}
	\label{tab:keypoints}
\end{table}

\noindent
\textbf{Effects of the Number of Keypoints.}
In Table~\ref{tab:keypoints}, we investigate the effects of the number of keypoints for encoding the scene features. Table~\ref{tab:keypoints} shows that larger number of keypoints achieves better performance, and similar performance is observed when using more than 4,096 keypoints. 
Hence, to balance the performance and computation cost, we empirically choose to encode the whole scene to 4,096 keypoints for the Waymo Open dataset. 
The above experiments show that our method can effectively encode the whole scene to 4,096  keypoints while keeping similar performance with a large number of keypoints, which demonstrates the effectiveness of the keypoint feature encoding strategy of our proposed PV-RCNN/PV-RCNN++ frameworks.

\begin{table}
	\begin{center}
		\scalebox{0.9}{
			\begin{tabular}{c|ccc|c}
				\hline
				{RoI-grid Size} & Vehicle  & Pedestrian & Cyclist & Average \\
				\hline
				$8 \times 8 \times 8$ & 68.88 & 63.74 & 67.84 & 66.82 \\
				$7 \times 7 \times 7$ & 68.76 & 63.81 & 68.00 & 66.85 \\
				$6 \times 6 \times 6$ & 68.62 & 63.74 & 68.26 & 66.87 \\
				$5 \times 5 \times 5$ & 68.28 & 63.54 & 67.69 & 66.50 \\
				$4 \times 4 \times 4$ & 68.21 & 63.58 & 67.56 & 66.45 \\
				$3 \times 3 \times 3$ & 67.33 & 62.93 & 67.22 & 65.83 \\
				\hline
		\end{tabular}}
	\end{center}
	\caption{Effects of the grid size in RoI-grid pooling module. All experiments are based on our PV-RCNN++ framework with a center-based head for proposal generation.}
	\label{tab:exp_grid_size}
\end{table}

\noindent
\textbf{Effects of the Grid Size in RoI-grid Pooling.}~
Table~\ref{tab:exp_grid_size} shows the performance of adopting different RoI-grid sizes for RoI-grid pooling module. We can see that the performance increases along with the RoI-grid sizes from $3\times3 \times3$ to $6\times 6\times 6$, and the settings with larger RoI-grid sizes than $6\times 6\times 6$ achieve similar performance. 
Hence we finally adopt RoI-grid size $6\times 6\times 6$ for the RoI-grid pooling module. 
Moreover, from Table~\ref{tab:exp_grid_size} and Table~\ref{tab:exp_improvements}, we also notice that PV-RCNN++ with a much smaller RoI-grid size $4\times 4 \times 4$ (66.45\% in terms of mAPH@L2) can also outperform PV-RCNN with larger RoI-grid size $6\times 6\times 6$ (65.24\% in terms of mAPH@L2), which further validates the effectiveness of our proposed sectorized proposal-centric sampling strategy and the VectorPool aggregation module.

\begin{table}
	\begin{center}
		\setlength\tabcolsep{2pt}
		\scalebox{0.87}{
		\begin{tabular}{cc|c|c|ccc|c}
			\hline
			{FoV}  & \#Points & Method & \tabincell{c}{Frame\\Rate} & Veh.  & Ped. & Cyc. & Average \\
			\hline 
			90$^{\circ}$ & $\sim$45k & PV-RCNN & 10.5 & 66.50 & 59.33 & 65.96 & 63.93\\
			90$^{\circ}$ & $\sim$45k & PV-RCNN++ & 16.0 & \textbf{67.29} & \textbf{61.51} & \textbf{66.91} & \textbf{65.24} \\
			\hline 
			180$^{\circ}$ & $\sim$90k & PV-RCNN & 6.8 & 65.34 & 60.20 & 61.70 & 62.41 \\
			180$^{\circ}$ & $\sim$90k & PV-RCNN++ & 12.3 & \textbf{66.05} & \textbf{62.08} & \textbf{63.44} & \textbf{63.86} \\
			\hline 
			360$^{\circ}$ & $\sim$180k & PV-RCNN & 3.3 & 67.54 & 61.62 & 66.57 & 65.24 \\
			360$^{\circ}$ & $\sim$180k & PV-RCNN++ & 10.0 & \textbf{68.62} & \textbf{63.74} & \textbf{68.26} & \textbf{66.87}\\
			\hline
		\end{tabular}
	}
	\end{center}
	\caption{Comparison of PV-RCNN/PV-RCNN++ on different sizes of scenes. ``FoV'' indicates the field of view of each scene, where for each scene in Waymo Open Dataset, we crop a specific angle (\emph{e.g.}, 90$^{\circ}$, 180$^{\circ}$) of frontal view for training and testing, and 360$^{\circ}$ FoV indicates the original scene. ``\#Points'' indicates the average number of points in each scene. ``Frame Rate'' indicates frames per seconds in terms of testing speed.}
	\label{tab:exp_scene_size}
	\vspace{-5mm}
\end{table}

\noindent
\textbf{Comparison on Different Sizes of Scenes.}~
To investigate the effects of our proposed PV-RCNN++ on handling large-scale scenes, we further conduct ablation experiments to compare the effectiveness and efficiency of PV-RCNN and PV-RCNN++ frameworks on different sizes of scenes. As shown in Table~\ref{tab:exp_scene_size}, we compare these two frameworks on three sizes of scenes by cropping different angles of frontal view of the scene in Waymo Open Dataset for training and testing.
PV-RCNN++ framework consistently outperforms previous PV-RCNN framework on all three sizes of scenes with large performance gains. 
Table~\ref{tab:exp_scene_size} also demonstrates that as the scales of the scenes get larger, PV-RCNN++ becomes much more efficient than PV-RCNN. 
In particular, when the comparison is conducted on the original scene of Waymo Open Dataset, the running speed of PV-RCNN++ is about three times  faster than PV-RCNN, demonstrating the efficiency of PV-RCNN++ on handling large-scale scenes.

\section{Conclusion}
In this paper, we present two novel frameworks, named PV-RCNN and PV-RCNN++, for accurate 3D object detection from point clouds. 
Our PV-RCNN framework adopts a novel voxel set abstraction module to deeply integrates both the multi-scale 3D voxel CNN features and the PointNet-based features to a small set of keypoints, and the learned discriminative keypoint features are then aggregated to the RoI-grid points through our proposed RoI-grid pooling module to capture much richer contextual information for proposal refinement. 
Our PV-RCNN++ further improves PV-RCNN framework by efficiently generating more representative keypoints with our novel sectorized proposal-centric keypoint sampling strategy, and also by equipping with our proposed VectorPool aggregation module to learn structure-preserved local features in both the voxel set abstraction module and RoI-grid pooling module. Thus, our PV-RCNN++ finally achieves better performance with much faster running speed than the original PV-RCNN framework.

Our final PV-RCNN++ framework significantly outperforms previous 3D detection methods and achieve state-of-the-art performance on both the validation set and testing set of the large-scale Waymo Open Dataset, and extensive experiments have been designed and conducted to deeply investigate the individual components of our proposed frameworks.

\bibliographystyle{spbasic}      
\bibliography{egbib_simple}   

\end{document}